\pdfoutput=1

\documentclass[11pt]{article}

\usepackage[preprint]{acl}

\usepackage{amsmath}  
\usepackage{esint}     

\usepackage{times}
\usepackage{latexsym}

\usepackage{booktabs} 
\usepackage{graphicx} 

\usepackage[T1]{fontenc}

\usepackage[utf8]{inputenc}

\usepackage{microtype}

\usepackage{inconsolata}

\usepackage{graphicx}

\usepackage{hyperref}       
\usepackage{url}            
\usepackage{booktabs}       
\usepackage{amsfonts}       
\usepackage{nicefrac}       
\usepackage{microtype}      
\usepackage{xcolor}         
\usepackage{comment}
\usepackage{tcolorbox}
\usepackage{xspace}

\usepackage{longtable}
\usepackage{booktabs} 
\usepackage{caption}  

\usepackage{orcidlink}
\usepackage{comment}
\usepackage{wasysym}
\usepackage{listings}
\usepackage{xcolor}
\usepackage{float}
\usepackage{graphicx}
\usepackage{epstopdf}
\usepackage{booktabs}
\usepackage{wrapfig}
\usepackage{amsmath} 
\usepackage{wasysym}
\usepackage{algorithm}
\usepackage{algpseudocode}
\usepackage{enumitem} 

\definecolor{codegreen}{rgb}{0,0.6,0}
\definecolor{codegray}{rgb}{0.5,0.5,0.5}
\definecolor{codepurple}{rgb}{0.58,0,0.82}
\definecolor{backcolour}{rgb}{0.95,0.95,0.92}
\lstdefinestyle{mystyle}{
    backgroundcolor=\color{backcolour},   
    commentstyle=\color{codegreen},
    keywordstyle=\color{magenta},
    numberstyle=\tiny\color{codegray},
    stringstyle=\color{codepurple},
    basicstyle=\ttfamily\scriptsize,
    breakatwhitespace=false,         
    breaklines=true,                 
    captionpos=b,                    
    keepspaces=true,                 
    showspaces=false,                
    showstringspaces=false,
    showtabs=false,                  
    tabsize=2
}
\let\cite\citep

\title{Multimodal Interactive Contextualized Real World Task Assistance from a Single Demonstration}
\title{Multimodal Cues for Grounding Real World Task Assistance}
\title{Grounding Task Assistance with Multimodal Cues}
\title{Grounding Task Assistance with Multimodal Cues from a Single Demonstration}

\author{
    \textbf{Gabriel Sarch}\thanks{\ \ This work was conducted while interning at Microsoft Research. Correspondence: \texttt{gsarch@andrew.cmu.edu}.} \quad
    \textbf{Balasaravanan Thoravi Kumaravel} \quad
    \textbf{Sahithya Ravi} \quad
    \textbf{Vibhav Vineet} \quad
    \textbf{Andrew D. Wilson} \\
    \\
    Microsoft Research, USA \\
}

\newcommand{\SystemName}{\textsc{MICA}\xspace}

\begin{document}
\maketitle
\begin{abstract}
A person's demonstration often serves as a key reference for others learning the same task.  However, RGB video, the dominant medium for representing these demonstrations, often fails to capture fine-grained contextual cues such as intent, safety-critical environmental factors, and subtle preferences embedded in human behavior. This sensory gap fundamentally limits the ability of Vision Language Models (VLMs) to reason about \emph{why} actions occur and \emph{how} they should adapt to individual users. To address this, we introduce \textbf{MICA (Multimodal Interactive Contextualized Assistance)}, a framework that improves conversational agents for task assistance by integrating eye gaze and speech cues. MICA segments demonstrations into meaningful sub-tasks and extracts keyframes and captions that capture fine-grained intent and user-specific cues, enabling richer contextual grounding for visual question answering. Evaluations on questions derived from real-time chat-assisted task replication show that multimodal cues significantly improve response quality over frame-based retrieval. Notably, gaze cues alone achieves 93\% of speech performance, and their combination yields the highest accuracy. Task type determines the effectiveness of implicit (gaze) vs. explicit (speech) cues, underscoring the need for adaptable multimodal models. These results highlight the limitations of frame-based context and demonstrate the value of multimodal signals for real-world AI task assistance.
\end{abstract}

\begin{figure}
  \includegraphics[width=0.8\columnwidth]{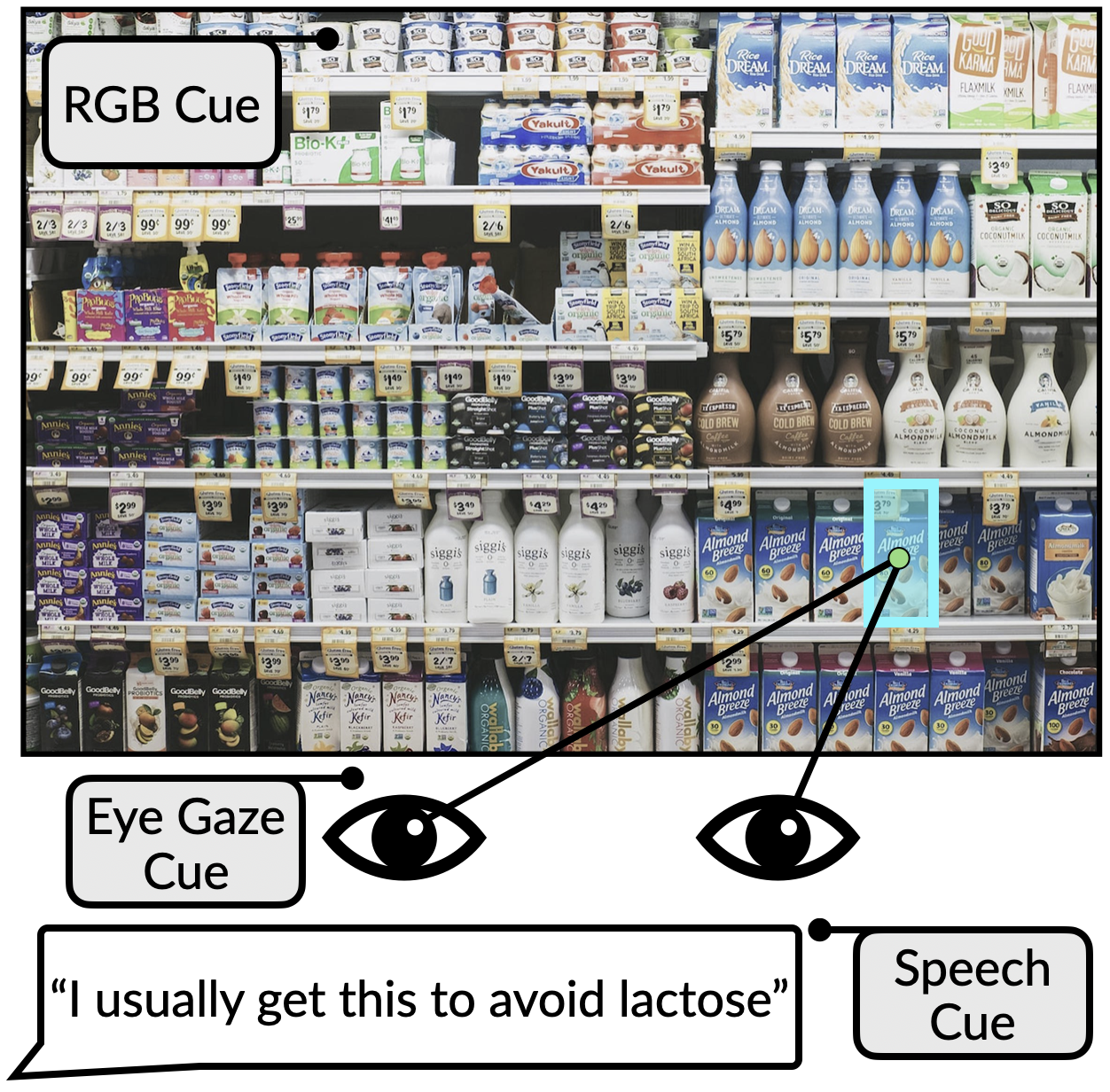}
  \caption{Illustration of multimodal cues in shopping. A user selects a lactose-free product using \textit{eye gaze} and \textit{speech}. Gaze highlights the product before speech clarifies intent: \textit{``I usually get this to avoid lactose.''}
}
\vspace{-1em}
  \label{fig:actually_teaser}
\end{figure}

\begin{figure*}
  \centering\includegraphics[width=1.0\textwidth]{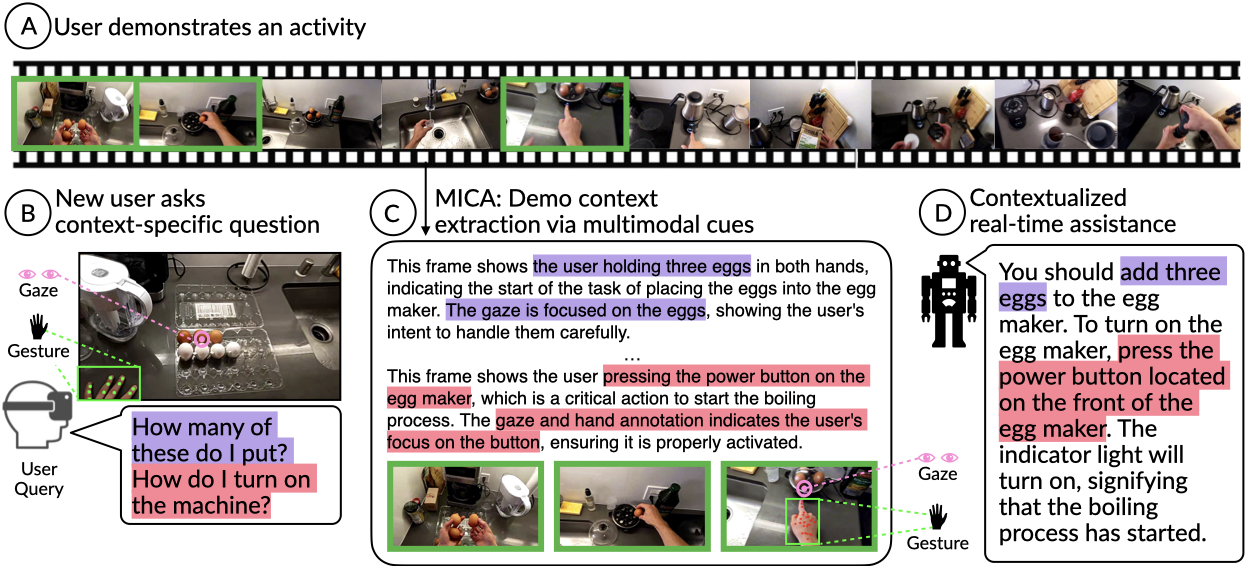}
  \caption{
  Overview of MICA (Multimodal Interactive Contextualized Assistance).
  (A) A user demonstrates an activity (e.g., boiling eggs) with multimodal inputs like RGB, speech, and gaze, with key frames highlighted in green. (B) A new user asks context-specific questions (purple: objects, red: actions). (C) MICA extracts contextual cues (e.g., egg count in purple, actions in red) using multimodal signals. (D) MICA provides real-time, personalized instructions, aligning object (purple) and action (red) references with the query. 
}
  \label{fig:teaser}
\end{figure*}

\section{Introduction}

When individuals learn tasks through demonstrations, whether administering medication, assembling machinery, or cooking family recipes, they rely on subtle contextual cues. Although pre-trained vision language models (VLMs) have shown promise in describing RGB video content, they often fail to ground the intent and fine-grained visual details in the demonstrations: a muttered safety warning, 
shifting gaze between relevant items or deliberately avoiding them,
or hesitant gestures revealing personal preferences. As illustrated in Figure~\ref{fig:actually_teaser}, both implicit (eye gaze) and explicit (speech) cues play a role in user intent and task execution, yet frame-based models often ignore these signals~\citep{10.1145/3332165.3347933}. This gap is especially problematic when safety requirements, implicit reasoning, or fine-grained preferences are essential, issues that generic training data do not capture.

The core challenge lies in reconciling two shortcomings: \emph{(1) partial perception}, where frame-only processing discards intent-revealing signals like spoken instructions and gaze identification, and \emph{(2) static reasoning}, where VLMs lack mechanisms to adapt to demonstrator-specific patterns. However, most retrieval-augmented generation (RAG) approaches limit themselves to captioning frames or clustering segments~\citep{wang2024videotree,videoagent}, losing these essential signals. 
Meanwhile, existing VLMs often struggle to ground fine-grained spatial details in images or video, particularly in the absence of additional guidance~\citep{fu2024blink,wu2024v,yang2024think}.
Consequently, critical moments, such as a chef's glance at oven settings or a mechanic's triple tool check, become lost as actionable signals.

To bridge this gap, we introduce \textbf{MICA (Multimodal Interactive Contextualized Assistance)}, an end-to-end system that records and processes a single user demonstration—including eye gaze and speech—and subsequently provides personalized guidance to a different individual. As shown in Figure~\ref{fig:teaser}, 
MICA extracts key moments from demonstrations and aligns them with user queries to provide contextualized assistance. MICA dynamically leverages available cues, using gaze and speech signals to segment the demonstration, extract key frames, and generate contextualized captions for retrieval-augmented generation. 
At inference time, it retrieves the most relevant segments and prompts a vision-language model on the contextualized images and captions to produce tailored responses.

We evaluate MICA on a benchmark derived from real-time chat-assisted interactions. In this setting, participants demonstrate personalized tasks while simultaneously providing multimodal cues. Later, \emph{different} users ask questions about these tasks to an AI chat assistant, aiming to replicate the same procedures. 
While previous video benchmarks rely solely on RGB and use offline annotators or templates for question generation~\citep{fu2024video,yang2024think}, ours integrates video, continuous gaze, and speech with ecologically-valid, on-the-fly user queries.
Our findings reveal: (1) MICA \textbf{significantly outperforms} zero-shot and frame-only RAG baselines, leveraging multimodal cues for richer and more contextualized responses. (2) GPT-4o effectively utilizes gaze to approach speech performance, \textbf{achieving 93\% of the speech condition’s score}. Furthermore, combining gaze and speech yields the highest performance. (3) Task type interacts significantly with cue effectiveness, showing that implicit (gaze) versus explicit (speech) cues vary in importance depending on the task. Certain tasks benefit more from implicit gaze cues (e.g., identifying preferences in shopping), while others rely on explicit verbalization (e.g., step-by-step routines). (4) Our in-depth analysis of the real-time dataset reveals that demonstration utterances and interaction questions are both task-dependent and goal-driven—factors often overlooked by existing benchmarks.


By unifying speech with RGB and nonverbal cues like gaze, MICA improves intent-aware task understanding—showing that \emph{how} demonstrators communicate (through hesitations and attention shifts) is every bit as crucial as \emph{what} they say.


\section{Related Works}
\label{sec:relatedworks}

\subsection{Multimodal Interaction and Gaze-Based Input}

Gaze has long been studied as an interaction modality~\cite{Jacob1990}, enabling hands-free interaction in graphical user interfaces. While early work focused on explicit gaze input (e.g., gaze-based typing), later approaches leveraged implicit gaze cues to infer user attention~\cite{Duchowski2004}. Understanding gaze behavior has been key to optimizing human-computer interaction, including eye-hand coordination~\cite{binsted2001eye,Barton2000} and active sensing in virtual environments~\cite{Sitzmann2018}. 

Recent research has shifted toward egocentric eye-tracking, particularly in augmented reality (AR) systems~\cite{Plopski2022}. With AR headsets like Microsoft HoloLens 2 and Apple Vision Pro, gaze tracking enables object selection and interaction in immersive environments. More broadly, multimodal interaction integrates gaze, gestures, and voice for richer user experiences. Early work, such as Bolt’s “Put-That-There”~\cite{Bolt1980}, demonstrated the synergy between speech and gestures, while later systems~\cite{koons1991integrating,cohen1997quickset} advanced multimodal fusion. Gaze enhances reference resolution in speech-based interactions~\cite{prasov2008s,Lee2024}, and recent studies~\cite{Khan2022,Bader2009} explore gaze-speech integration for implicit interactions.

\subsection{Egocentric Task Assistance and Guidance}

Egocentric video QA systems, such as Ego4D~\cite{grauman2022ego4d} and Ego-Exo4D~\cite{grauman2024ego}, retrieve episodic memory from extensive first-person video datasets but primarily focus on passive, retrospective analysis rather than real-time task assistance. In contrast, our approach leverages a single-task demonstration to extract multimodal cues—gaze, speech, and gestures—enabling real-time, interactive guidance.

Automated task guidance systems provide step-by-step assistance using sensor-equipped environments~\cite{schoop2016drill,SmartMakerspace} or AR-based tracking~\cite{DuploTrack, 10.1145/3332165.3347872}, though many rely on extensive user-defined tutorials and lack adaptability across different settings~\cite{whitlock2020authar,ProcessAR,huang2021adaptutar}. Recent AR-based assistants, such as SIGMA~\cite{Bohus2024,bohus2024sigmaopensourceinteractivemixedreality}, integrate generative AI for personalized task guidance, while systems like GazePointAR~\cite{Lee2024} and GazeGPT~\cite{konrad2024gazegptaugmentinghumancapabilities} incorporate gaze for adaptive interaction. Our approach builds on these advances by learning from demonstrations without requiring manual authoring, improving generalization across diverse tasks.

\subsection{Visual Retrieval-Augmented Generation (RAG)}

Visual RAG models offer an alternative to traditional fine-tuned models by retrieving relevant visual data in real time~\cite{majumdar2024openeqa,wang2023holoassist}. In-context learning has been explored in text~\cite{brown2020language,agarwal2024many}, image~\cite{bar2022visual,wang2023images}, and image-text domains~\cite{zheng2024large}, though performance depends on selecting relevant context examples~\cite{zhang2023makes,balazevic2024towards}. Unlike prior work that primarily relies on image features, our multimodal RAG framework integrates explicit and implicit cues, optimizing retrieval for personalized assistance tasks. This allows for adaptive, real-time guidance without requiring large-scale dataset-specific training.

\begin{figure*}[htbp]
    \centering
    \includegraphics[width=1.0\textwidth]{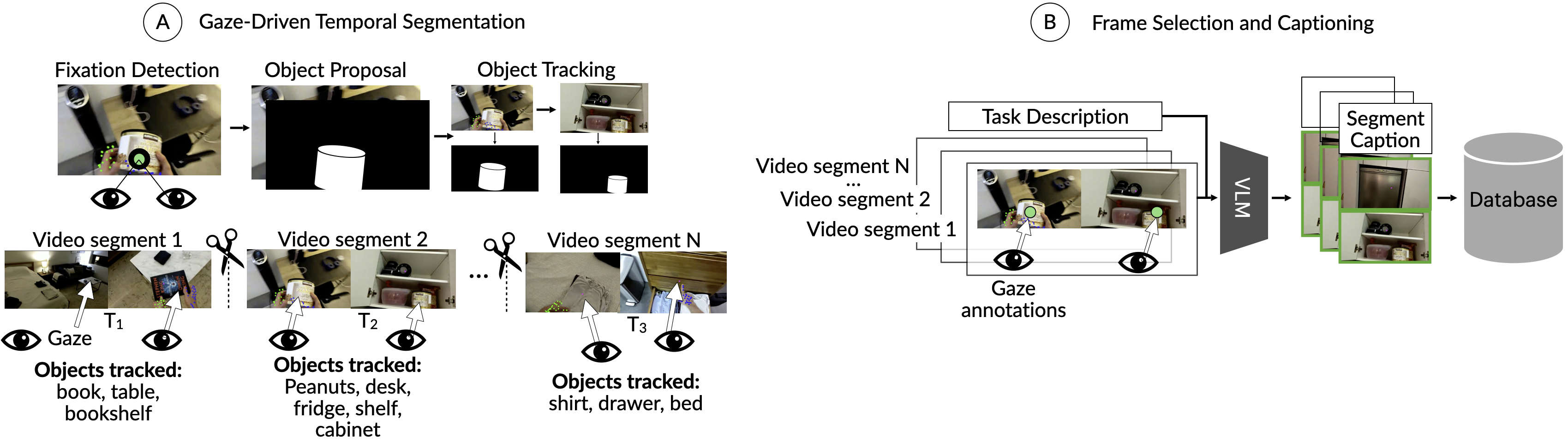}
    \caption{\textbf{MICA uses multimodal cues to extract context from demonstrations.} The framework leverages gaze-based cues to segment the demonstration temporally (A) For gaze-driven temporal segmentation, we monitor changes in the user’s gaze to detect and track gaze-fixated objects. Once the set of fixated objects changes substantially, a new temporal segment is started. (B) For each temporal segment, key frames and captions are generated via a Vision-Language Model by conditioning on gaze and/or speech annotations. The resulting information is saved to a database for retrieval-augmented assistance.}
    \label{fig:multimodal_cues}
\end{figure*}

 
    

\section{Problem Formulation}
\label{problem_form}
Consider the scenario in which a conversational assistant helps a new user replicate an everyday task after seeing a single demonstration from a different user. Formally, given a query \((Q, I)\) from a new user which consists of an open-vocabulary question \(Q\) and an image \(I\) (Fig \ref{fig:teaser}B). The goal of the assistant is to generate an answer \(A\) (Fig \ref{fig:teaser}D) to the question \(Q\) informed by the image \(I\) and a previous demonstration \(D\) (Fig \ref{fig:teaser}A). The answer  \(A\) must approximate the ground truth human answer \(A^{*}\) i.e, \(A \approx A^{*}\).


The demonstration \(D\) includes:
1. Egocentric RGB frames \(I_t\) at each time step $t$.
2. Natural language speech transcriptions \(N = (n, s, e)\), where $n$ represents the text segment spoken by the user, $s$ denotes the start time, and $e$ denotes the end time.
3. Eye gaze \(g_t = (p_t, d_t)\), with gaze origin $p_t \in \mathbb{R}^3$ and gaze direction $d_t \in \mathbb{R}^3$, and hand pose \(K_t\) $\in \mathbb{R}^{3 \times k}$ for each hand keypoint $k$. Despite overlaying hand pose keypoints on all frames, they had little effect on interpreting hand cues; we therefore prioritize gaze and speech. 


\section{Method} \label{sec:methods}
The goal of our framework shown in Fig.~\ref{fig:teaser}C is to incorporate implicit (eye gaze) and explicit (speech) cues to enrich demonstration understanding for Vision-Language Models (VLMs). As shown in (Fig.~\ref{fig:multimodal_cues}), we first segment demonstrations into meaningful units and then perform key frame extraction and captioning. 


\subsection{Gaze-based Temporal Segmentation}

To structure the demonstration into meaningful units, we segment a long demonstration based on user actions or ``key moments.'' This segmentation ensures that subsequent processing steps operate on coherent, task-relevant segments rather than raw, unstructured video frames.  


We segment the video based on changes in the user’s visual attention, as shown in Figure~\ref{fig:multimodal_cues}A. At each frame, we generate object proposals at the user's gaze point using SAM~\citep{kirillov2023segment}. We track these objects over time with DEVA~\citep{cheng2023tracking} to maintain continuity. A segment boundary is defined when the set of tracked objects changes by a fixed percent (i.e., fixation shifts to a different object or group of objects for a sustained period). This ensures segments reflect meaningful task transitions rather than brief gaze shifts. Further implementation details are in the Appendix Section~\ref{temp_seg}.

\subsection{Key Frame Extraction and Captioning}
Once the video is segmented, we extract representative frames and generate textual descriptions for each segment, as shown in Figure~\ref{fig:multimodal_cues}B. This process provides a structured summary that enables efficient retrieval at inference time. For each segment, we uniformly sample 30 frames from the segment and prompt a VLM to:
\begin{enumerate}
    \item Select the top-\(k\) key frames that best represent the segment. We use $k=3$.
    \vspace{-1em}
    \item Generate detailed captions describing the user’s activity and context within the segment.
\end{enumerate}
Gaze-based methods overlay a projected gaze point onto each frame. Speech-based methods append the user’s utterances to the prompt. We give an overall task description to the VLM (e.g., ``The user is shopping''), which may be inferred or provided. The resulting key frames and captions form a database entry for retrieval-augmented generation at inference time.

\subsection{Inference on New Questions}
\label{sec:inference}
When a new user asks a question with an accompanying image (e.g., from their wearable device), we:

\begin{enumerate}
    \item Caption the query image with a captioning model.
    \vspace{-0.75em}
    \item Encode both the caption and image into embeddings.
    \vspace{-0.75em}
    \item Retrieve the top-\(k\) matching segments from the stored database.
    \vspace{-0.75em}
    \item Append these retrieved segments (key frames and captions) into the VLM’s context window.
\end{enumerate}

Finally, the VLM produces an answer. Implementation details, including the similarity scoring function and weighting of textual vs.\ visual embeddings, can be found in the Appendix Section~\ref{inference}.

\section{Data Collection}

\label{sec:data}
\begin{figure}[H]
    \centering
    \includegraphics[width=1.0\columnwidth]{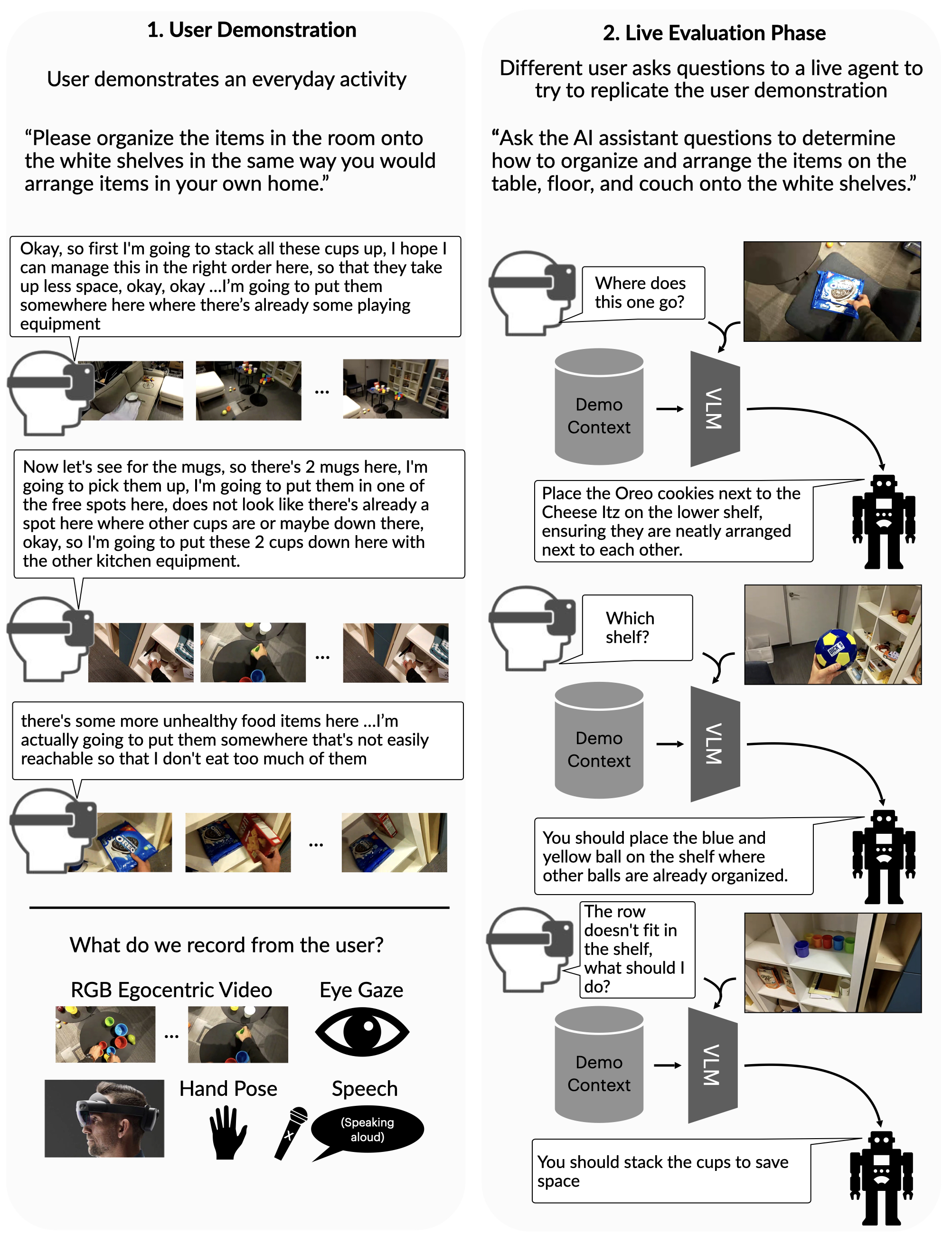}
    \caption{In the demonstration phase (left panels), a user performs a task while wearing a HoloLens 2, recording their eye gaze and speech. During the live evaluation phase (right panels), the system uses the demonstration context to assist a new user in replicating the task by answering real-time questions.}
    \label{fig:system_plot}
\end{figure}

We collected user demonstrations and evaluations with a HoloLens 2 across three task categories: \emph{organizing a room, shopping, and a morning routine.} In the \textbf{demonstration phase} (Fig.~\ref{fig:system_plot}, left), participants performed each task following their own preferences while speaking aloud to provide rich verbal cues. In the \textbf{live evaluation phase} (Fig.~\ref{fig:system_plot}, right), a new participant asked questions to replicate the same task setup. This produced realistic, user-specific queries answered by referencing the demonstration. 

Our offline evaluation set includes 32 demonstrations and 415 live questions. We annotated each question with a ground-truth answer \(A^{*}\). Question accuracy is measured via LLM-Match (See Appendix Section~\ref{sec:llmmatch}). Further details on participant instructions, hardware setup, and sample prompts are provided in Appendix Sectoion~\ref{supp_data_collection}.

\section{Results}\label{sec:results}
We evaluate the effectiveness of various models in extracting user-specific context from demonstrations, leveraging implicit (e.g., eye gaze) and explicit (e.g., speech) cues. Our goal is to benchmark advanced VLMs' performance in understanding nuanced user intent and facilitating accurate responses in real-world applications.

\paragraph{Baselines.} 
We evaluate the ability of various baselines to extract relevant context from demonstrations using implicit and explicit cues on the annotated evaluation set.

For each model, we assess the following methods for in-context learning from the demonstration:

1. \textsc{Zero Shot:} The model is provided with the query $Q$ without any additional context from the demonstration.

2. \textsc{CLIP Clustering:} A visual clustering approach that groups similar frames based on feature similarity derived from a pre-trained feature encoder, CLIP~\citep{radford2021learning}. This method is commonly used in retrieval-based video understanding techniques~\cite{wang2024videotree}. We encode all video frames and apply k-means clustering with $k=10$. For each cluster, we select the top-3 keyframes closest to the cluster centroid vector and provide them to GPT4o for captioning, as detailed in Section~\ref{key_frame}. For a given question, retrieval occurs the in the same manner as defined in Section~\ref{inference}. 

3. \textsc{Frames as Context:} Frames are encoded as CLIP embeddings, and for each input, the 10 frames with the lowest L2 distance to the input query frame are retrieved to serve as context, as described in Section~\ref{inference}.

\paragraph{MICA Condition Comparisons.} We test our method, \SystemName, with different cues available during demonstration processing:

\hangindent=4mm 1. \textsc{MICA Eye Gaze:} Utilizes eye gaze cues without speech inputs, as detailed in Section~\ref{sec:methods}. 

\hangindent=4mm 2. \textsc{MICA Speech:} Leverages speech cues without gaze. Instead of gaze-based temporal segmentation, we use speech transcriptions from Whisper to identify segment boundaries. Each utterance (with start and end times) becomes one segment.

\hangindent=4mm 3. \textsc{MICA Eye Gaze + Speech:} As described in Section~\ref{sec:methods}, combines eye gaze for segment generation with speech transcripts in prompts to select keyframes and caption video segments.

\hangindent=4mm 4. \textsc{MICA Eye Gaze + Speech + Summary:} Extends the Eye Gaze + Speech method by appending an inferred textual summary of the demonstration generated by prompting the VLM with all segments.

Unless the method name explicitly includes "w/ Inferred Intent," all reported results use ground-truth task descriptions (e.g., "The user is cleaning a room").


\paragraph{Models.} We evaluate each method using the following models: GPT4o, GPT4o-mini, VILA-1.5-3B with context provided by GPT4o (which uses a database of text and images obtained using GPT4o), LLaVA-OneVision-7B with GPT4o context (also utilizing a database of text and images from GPT4o), and VILA-1.5-3B independently.


We investigate the following research questions:

\begin{enumerate}[label=\large\protect\textcircled{\small\arabic*}]
    \item\textbf{RQ1}: How well can current VLMs utilize implicit and explicit cues for user-specific context understanding?
    \item \textbf{RQ2}: What is the effect of task type on implicit and explicit context understanding?
    \item\textbf{RQ3}: What verbal behaviors do participants exhibit while demonstrating tasks and interacting with a live AI assistant?
\end{enumerate}

\begin{figure*}[htbp]
    \centering
    \includegraphics[width=0.8\textwidth]{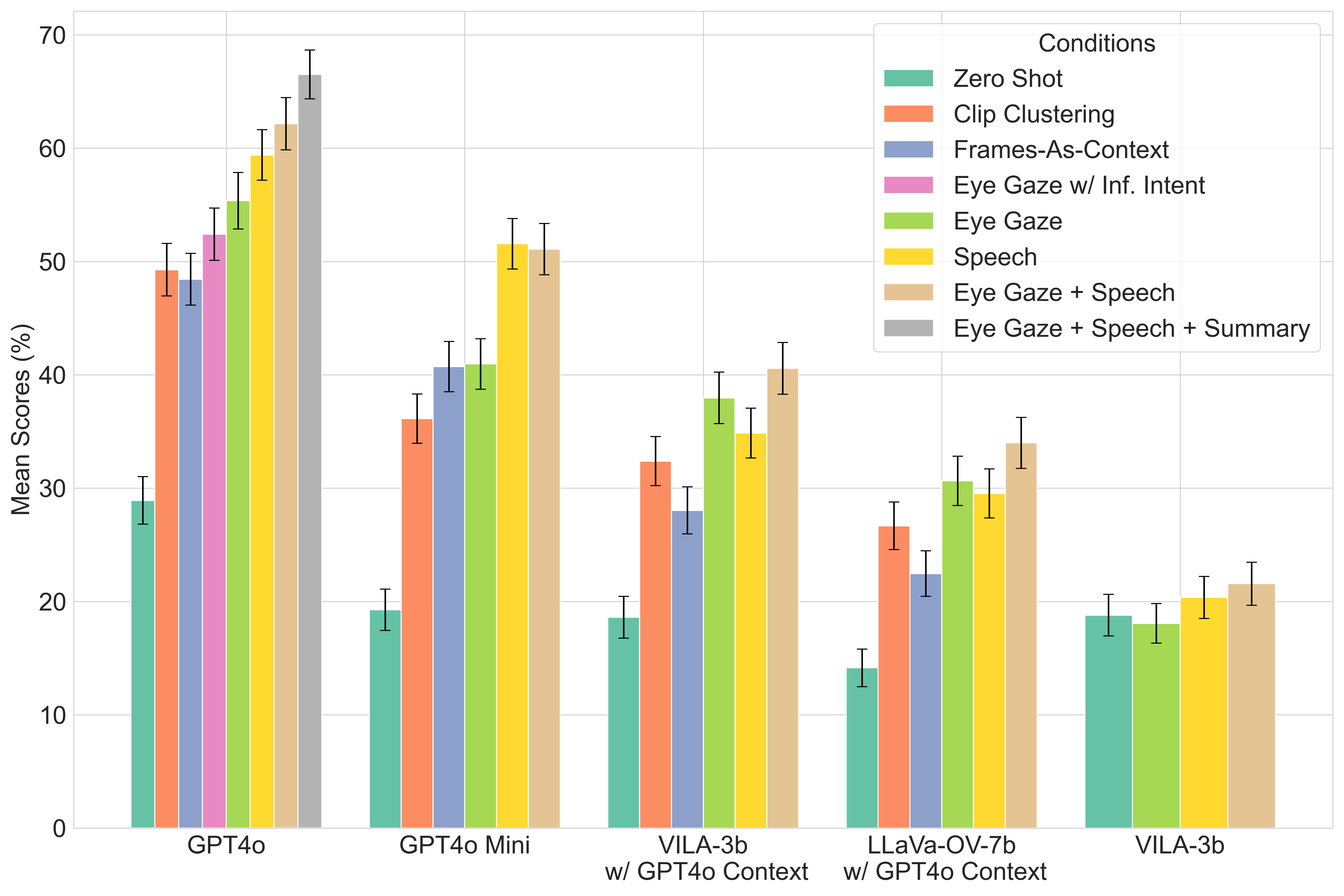}
    \caption{
    Performance of base models (GPT4o, GPT4o-mini, VILA, LLaVA) across different context extraction methods (Zero Shot, CLIP Clustering, Frames as Context, Eye Gaze, Speech, Eye Gaze + Speech) on the annotated evaluation set. 
    The plot shows overall model performance across questions.
    We report mean $\pm$ standard error across evaluation questions (n=415). We include a table version of this plot in Appendix Table~\ref{tab:condition_comparison}.
    }
    \label{fig:gazevspeech}
\end{figure*}

\subsection{RQ1: How well can current VLMs utilize implicit and explicit cues for user-specific context understanding?}\label{sec:Q1}

Our main findings are shown in Figure~\ref{fig:gazevspeech}. We draw the following conclusions.

\textbf{1. Zero-shot baselines significantly underperform.} Pretrained knowledge alone is insufficient to respond to user queries accurately, emphasizing the necessity of demonstration context.

\textbf{2. \SystemName{} outperforms frame-only RAG baselines} 
Across all models, MICA consistently outperforms frame-only RAG baselines. Using GPT-4o, the \textsc{Frames-as-Context} baseline achieves 48.4\% $\pm$ 2.1 (Figure 5; GPT4o, \textsc{Frames-as-Context}) while even the most minimal MICA variant (\textsc{Eye Gaze}) already improves upon this at 55.37\% $\pm$ 2.3 (Figure 5; GPT4o, \textsc{Eye Gaze}). Adding speech further improves results: \textsc{Eye Gaze + Speech} achieves 62.2\% $\pm$ 2.3 (Figure 5, GPT-4o, \textsc{Eye Gaze + Speech}), and \textsc{Eye Gaze + Speech + Summary} reaches the highest performance at 66.5\% $\pm$ 2.2 (Figure 5, GPT-4o, \textsc{Eye Gaze + Speech + Summary}). This progression highlights MICA’s strength in integrating multiple modalities to produce more contextually grounded responses above frame-only cues.

\textbf{3. GPT4o utilizes gaze input to approach speech performance (93\%).} With a high-level task description, GPT4o demonstrates gaze as a viable alternative to speech (55.3\%$\pm$2.4 vs. 59.4\%$\pm$2.5; U test n.s., p=0.25). Combining gaze and speech yields the highest performance (62.16$\pm$2.30), and adding an inferred demonstration summary further improves scores by 4.35\% (66.51$\pm$2.15). 

\textbf{4. Inferring task descriptions degrades gaze performance but not speech.} 
With GPT-4o, \textsc{Eye Gaze} (with ground-truth intent) reaches 55.4\%$\pm$2.5 (Figure 5, GPT-4o, \textsc{Eye Gaze}), but performance declines to 52.4\%$\pm$2.3 (Figure 5, GPT-4o, *Eye Gaze w/ Inf. Intent*) under the *Eye Gaze w/ Inferred Intent* condition—a decrease of 3.0\%, consistent across tasks. In contrast, the \textsc{Speech} condition remains stable: 59.4\%$\pm$2.2 with ground-truth intent (Figure 5, GPT-4o, \textsc{Speech}) versus 59.4\%$\pm$2.15 with inferred intent (result not shown in the figure). Thus, speech cues that include the user's explicit statement of intent offer a clearer and more reliable signal for inferring intent (U-test p = 0.033) and supporting task-level abstraction.

\textbf{5. Weaker VLMs struggle with gaze input.} GPT4o-mini significantly underperforms in gaze conditions (40.9\%$\pm$2.2 vs. 51.6\%$\pm$2.2; U test, p=0.0007), suggesting an inability to infer context from implicit cues. 

\textbf{6. GPT4o enhances weaker models’ performance.} When GPT4o processes context for weaker models (e.g., VILA, LLaVA), these models improve significantly, achieving parity between gaze and speech conditions (e.g., VILA-1.5-3b; 37.9\%$\pm$2.3 vs. 34.8\%$\pm$2.2; U test, p=0.38). 

\textbf{7. Small open-source models struggle with video context extraction.} Compared to closed-source VLMs, VILA 3B fails across all conditions, underscoring its limitations in understanding user intent in videos.

\subsection{RQ2: What is the effect of task type on implicit and explicit context understanding?}

\begin{figure}[t]
    \centering
    \includegraphics[width=1.0\columnwidth]{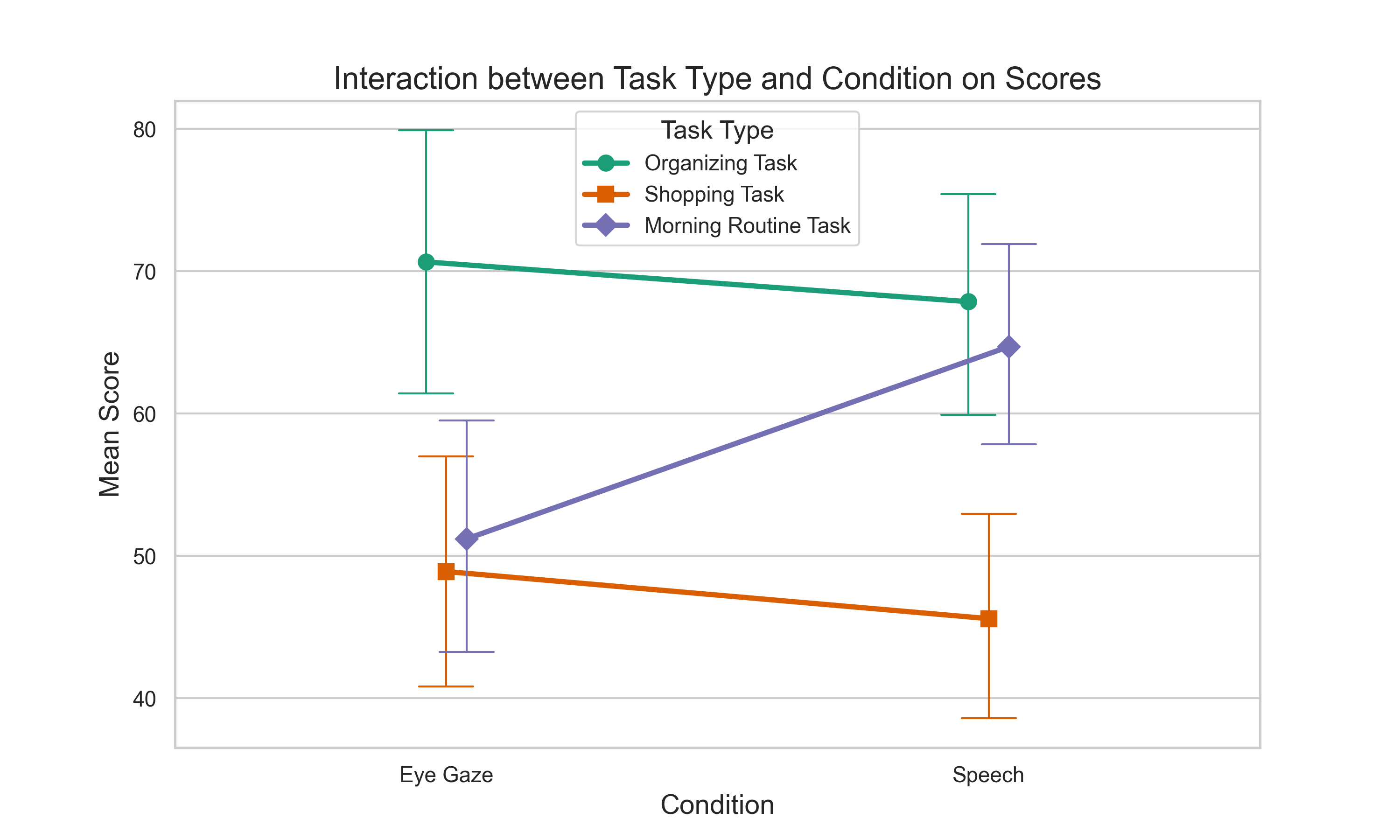}
        \caption{Interaction effect between task type and implicit (eye gaze) or explicit (speech) condition. We report the mean score and 95\% CI for each task type and condition. A significant interaction between task type and condition ($\beta = 0.3485$, $p = 0.049$) highlights differential impacts of Eye Gaze and Speech across tasks, with Morning Routine improving under Speech while Organizing and Shopping tasks show varying changes. A significant main effect of task type ($\beta = -0.3904$, $p = 0.004$) suggests scores decrease from Organizing to Shopping to Morning Routine, irrespective of condition.}
    \label{fig:task_effect}
\end{figure}

To determine the effect of task type on outcomes under implicit or explicit cues, we examined the effects of different task types (Organizing, Shopping, Morning Routine) on evaluation accuracy under two conditions (Eye Gaze, Speech) using GPT4o context extraction using an ordinal mixed-effects model. The model included fixed effects for task and condition, as well as their interaction, with random intercepts for conditions to account for variability between groups. 

As shown in Figure~\ref{fig:task_effect}, results from the ordinal mixed-effects model indicate a significant interaction between task type and condition ($\beta = 0.3485$, $p = 0.049$), suggesting that the effect of task type on scores is moderated by the condition (speech or gaze). Specifically, task types exhibited varying responses under Eye Gaze and Speech conditions, as depicted in Figure~\ref{fig:task_effect}. For instance, while the Morning Routine task showed improved scores in the Speech condition compared to Eye Gaze, the Organizing and Shopping tasks demonstrated either a decline or minimal change. This differential impact underscores the importance of considering both task type and context condition when evaluating task performance.

Furthermore, the main effect of task type was found to be statistically significant ($\beta = -0.3904$, $p = 0.004$), indicating that scores generally decrease with the extent of changes in task type (Organizing > Shopping > Morning Routine), regardless of condition. 

\textbf{These findings show that task type significantly interacts with condition, highlighting that the effectiveness of implicit cues (Eye Gaze) versus explicit cues (Speech) varies across tasks.} This variability suggests that selecting the appropriate cue type is crucial for optimizing task performance, as certain tasks benefit more from implicit cues, like eye gaze, while others are better supported by explicit cues, like speech.


\begin{figure}[htbp]
    \centering
    \includegraphics[width=1.0\columnwidth]{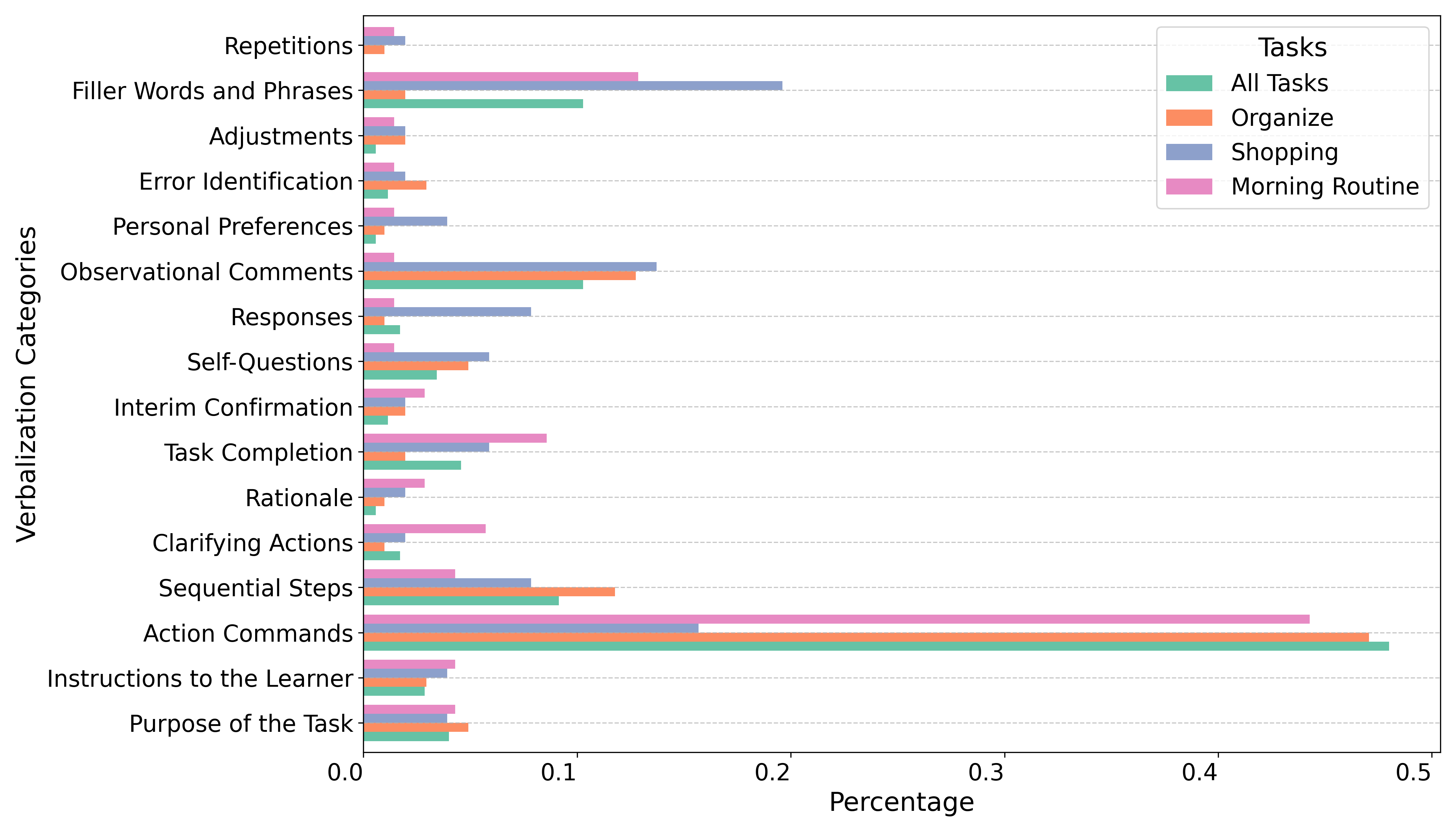}
    \caption{Distribution of instructional speech categories. \textit{Action Commands} dominate across tasks (48\%), with variations reflecting task-specific structures.}
    \label{fig:utterance_categories}
\end{figure}

\begin{figure}[htbp]
    \centering
    \includegraphics[width=1.0\columnwidth]{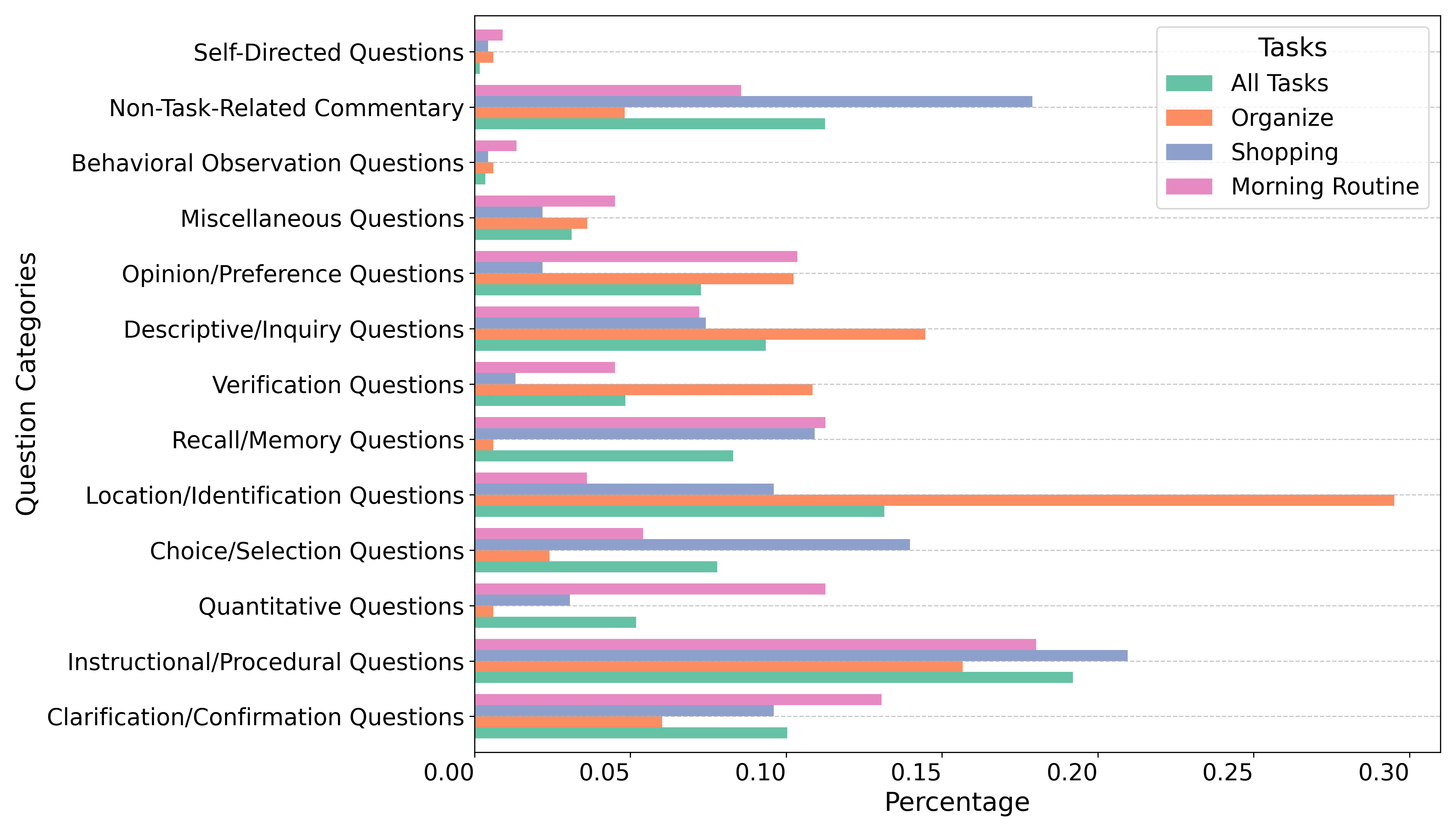}
    \caption{Distribution of question types. \textit{Instructional/Procedural Questions} dominate (19.2\%), with task-specific trends highlighting unique information needs.}
    \label{fig:question_categories}
\end{figure}

\subsection{RQ3: What verbal behaviors do participants exhibit while demonstrating tasks and interacting with a live AI assistant?}

While previous video benchmarks rely on offline annotators~\citep{fu2024video} or synthetic question generation~\citep{yang2024think}, our dataset uniquely captures live, interactive user behaviors during personalized demonstrations. This ecological approach provides deeper insight into the natural communication patterns users adopt when engaging directly with AI assistants.

We analyzed participants’ verbal behaviors, categorizing utterances during demonstrations using GPT-4o (Figure~\ref{fig:utterance_categories}). \textit{Action commands} dominated overall (48\%), indicating users primarily adopt directive speech to instruct AI systems clearly. Interestingly, task-specific differences emerged prominently: \textit{organizing} tasks showed structured speech, characterized by frequent sequential steps (11.8\%) and explicit error identification (2.9\%), demonstrating users' structured approach and goal-oriented clarity. In contrast, \textit{shopping} tasks exhibited fewer action commands (15.7\%) and increased conversational fillers (19.6\%), highlighting users' preference for a relaxed and reflective style indicative of ongoing decision-making. The \textit{morning routine} tasks notably involved more explicit task completion indicators (8.6\%) and clarifying utterances (5.7\%), reflecting user intent to demarcate completed activities clearly and seek reassurance in routine tasks.

In evaluating participant question types during live interactions (Figure~\ref{fig:question_categories}), we identified \textit{Instructional/Procedural Questions} as most prevalent (19.2\%), followed by \textit{Location/Identification} (13.15\%). \textit{Behavioral Observation Questions} appeared rarely (0.35\%), suggesting minimal expectation for AI visual awareness by current users. Task-specific question patterns further illustrate users’ adaptive strategies: \textit{organizing} predominantly involved queries about item placement (29.52\%), highlighting spatial reasoning demands; \textit{shopping} elicited frequent instructional (20.96\%) and choice-based questions (13.97\%), reinforcing continuous decision-making; and \textit{morning routines} featured frequent quantitative and recall inquiries (11.26\%), emphasizing the user's need to track task progression accurately.

\textbf{Overall, these findings underline that ecological interactions with AI assistants are highly nuanced, task-dependent, and driven by distinct information needs at different phases of task execution.} Effective AI-driven assistance must therefore dynamically adapt to the directive and evaluative verbal cues users naturally exhibit, providing timely guidance on next actions and situational object grounding. Future systems that integrate explicit instructions, clarifications, and conversational speech into their interaction paradigms could substantially improve user experience by better aligning AI behaviors with human expectations.
\section{Analysis of implicit cues to ground context}\label{sec:discussion}

Our results (Section~\ref{sec:Q1}) show that VLMs effectively use eye gaze for contextual understanding, matching or surpassing speech in some cases 
. Three key benefits emerged: (1) gaze provides precise focus cues for disambiguation, (2) it grounds intent by linking attention to objects, and (3) it resolves ambiguous references like `those’ more effectively than speech alone.

\paragraph{Item Identification and Selection.}
Eye gaze improves item recognition where speech is ambiguous. In one case, when asked, \textit{``Based on my previous shopping, what would you buy here?''}, the gaze model correctly identified ``green pear'' and ``red apple,'' leveraging extracted demonstration content (e.g., \textit{``places the red apple into a white bowl with the green pear''}). The speech-only model misidentified them as ``colorful toy fruits'' due to vague speech (e.g., \textit{``Ooh, they would probably love these''}).

Similarly, for \textit{``Which items did I buy last time to build the birdhouse?''}, gaze identified the correct tools (hammer, drill, tape measure) based on extracted cues (\textit{``picks up a hammer and places it into the crate''}). The speech-only model, relying on phrases like \textit{``I’m going to need this one''}, omitted key tools, leading to errors.


\paragraph{User Intent Clarification.}
Gaze-based models inferred user intent more effectively than speech-only models by incorporating non-verbal cues 

For \textit{``How many Oreos did I buy last time?''}, gaze correctly answered `one package’ based on segment details (\textit{``reaching for Oreo cookies and placing them in the basket''}), due to gaze focus on one package in the video. The speech model, lacking detailed visual grounding, incorrectly answered `two packages.'

\paragraph{Implicit Visual Referencing.}
Gaze enhances reference resolution, improving answer specificity 
. When asked, \textit{``Are you saying this one I should buy?''}, the gaze model disambiguated: \textit{``No, buy the yellow Vroom box''}. Without gaze, the model incorrectly confirmed a different toy due to requiring grounding of ``those''. This demonstrates how gaze improves referential grounding, allowing models to resolve ambiguous queries more effectively.

These results highlight the benefits of incorporating gaze cues in VLMs for contextual understanding, item recognition, and disambiguation, leading to more precise and context-aware AI assistance.



\section{Conclusion}
In this work, we presented a novel framework that leverages VLMs with gaze and speech inputs to provide contextualized task assistance from a single demonstration. Our approach highlights the potential of using both implicit and explicit cues to enable personalized and efficient task assistance in real-world settings. Through experiments across various tasks such as organizing, shopping, and morning routines, we demonstrated that VLMs can achieve significant performance gains compared to naive video methods by effectively utilizing implicit gaze cues.
\section{Limitations}\label{sec:lim_future}






\paragraph{System Limitations}

There are limitations in the current system design. For example, our approach primarily provides audio output, which limits the ways in which instructions and feedback can be delivered to users. Expanding to multimodal outputs, such as visual or holographic cues, could offer a richer user experience and potentially improve task performance.

Moreover, while we explored variations in object instances, users, and room arrangements between the demonstration and evaluation phases, the evaluation did not involve completely new environments between the demonstration and evaluation phases. Future research should explore how well the system generalizes to completely different environments and contexts, including variations in physical space.

Currently, the system is designed for single-task scenarios, where each database is tailored to a specific task. Expanding to multi-task learning and exploring how demonstrations of multiple tasks can be effectively integrated or combined will be a crucial next step. Additionally, while our system relies on gaze and speech as input modalities, future work could explore the integration of other inputs, such as ambient context, to provide a more comprehensive understanding of user intent.

\paragraph{Hardware and Performance Constraints}

Our system's performance is also influenced by its reliance on specific hardware, such as eye-tracking and speech recognition technologies. Although the approach is not tied to any specific AR device, certain hardware limitations, such as resolution and response latency, can impact overall performance. Notably, there is a tradeoff between the amount of context provided to the model and the response time in current VLMs; feeding excessive context can lead to significant latency, which may not be acceptable for time-sensitive tasks. Future improvements should focus on optimizing this balance to ensure both context richness and acceptable response times, particularly for assistive applications requiring rapid feedback.

\clearpage

\bibliography{sample-base,main}
\clearpage
\appendix
\makeatletter
\renewcommand \thesection{S\@arabic\c@section}
\renewcommand\thetable{S\@arabic\c@table}
\renewcommand \thefigure{S\@arabic\c@figure}
\makeatother

\setcounter{section}{0}
\setcounter{figure}{0}  
\setcounter{table}{0} 

\renewcommand{\theHsection}{Supplement.\thesection}
\renewcommand{\theHtable}{Supplement.\thetable}
\renewcommand{\theHfigure}{Supplement.\thefigure}


\section{Appendix Overview}
We organize the appendix as follows: 

\begin{itemize}
    \item \textbf{Section~\ref{risks}:} Discussion of potential risks.
    \item \textbf{Section~\ref{supp_methods}:} Additional details on methods.
    \item \textbf{Section~\ref{supp_data_collection}:} Further information on data collection.
    \item \textbf{Section~\ref{supp_add_results}:} Additional results and analysis.
    \item \textbf{Section~\ref{sec:utterance_categories}:} Details on behavioral coding.
\end{itemize}

\section{Potential Risks}\label{risks}
MICA relies on multimodal user data, including speech and gaze, which may raise privacy concerns if not handled securely. Additionally, biases in vision-language models could lead to misinterpretation of user intent, particularly for diverse user behaviors and accessibility needs. Future work should focus on robust privacy safeguards and bias mitigation strategies to ensure fair and responsible deployment.

\section{Additional Methods Details}\label{supp_methods}

\subsection{Inferring Overall User Intent}\label{infer_intent}

User intent refers to the underlying goal or purpose behind a user's actions, and accurately identifying it is crucial for providing personalized and contextually relevant assistance. In MICA, a ground truth user intent is provided obtained by annotators, but may be inferred by the model itself. To do so, given a demonstration $D$, we first infer the user's overall intent for the entire activity (e.g., ``The user is cleaning a room'') to contextualize all future steps towards the overall goal, as shown in Figure~\ref{fig:system_plot} A1 and A2. To fit within the VLM's context window, we uniformly sample 50 frames from the sequence $I$. These subsampled frames collectively represent the entire demonstration, and the VLM is prompted to infer and output a single, consistent intent that applies across all sampled frames, accurately describing the user's activity throughout the demonstration. To provide additional contextual cues for inferring this intent, along with these frames, we include annotations that consist of gaze data or speech data.

\begin{figure}
    \centering
    \includegraphics[width=0.6\columnwidth]{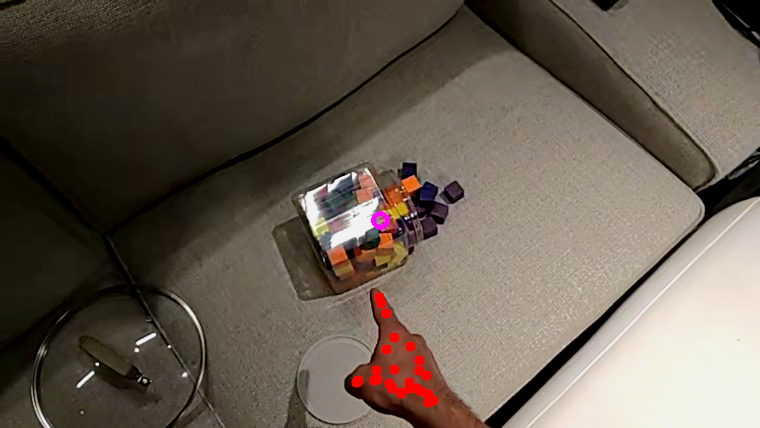}
    \caption{ Example of multimodal eye gaze and hand gesture overlay for the image input Vision-Language Model (VLM). MICA uses visual prompting to indicate gaze and gesture locations to the VLM. The user's gaze is represented by the purple circle, while hand gestures are highlighted in red.}
    \label{fig:gaze_overlay}
\end{figure}

For the gaze condition, we reproject the user's eye gaze from a 3D ray to a 2D image space using the camera's extrinsic and intrinsic parameters. We then use visual prompting~\cite{cai2023vipllava} to highlight the gaze point on the image and reference this point in the prompt. For the speech condition, we transcribe any speech uttered by the user during the demonstration and append these transcriptions to the prompt.  An example of this can be seen in Figure \ref{fig:gaze_overlay}.



\subsection{Temporal Segmentation}\label{temp_seg}

To accurately capture and understand the sequence of events—discrete actions or occurrences within the activity—in a long video demonstration, it is crucial to divide the video into distinct temporal segments.
By breaking the video down into smaller segments, we can better identify and track changes in user behavior and object interactions, leading to a more comprehensive understanding of the overall demonstration. As before, we consider 
a temporal segmentation method based on gaze or speech, as shown in Figure~\ref{fig:system_plot} B1 and B2.



\subsubsection{Eye-Gaze-Based Object Tracking and Temporal Segmentation}

We propose an approach to leverage eye gaze data to perform object tracking and key moment boundary detection in videos. This method involves three main steps: 1) detecting fixations using in-clip consensus, 2) generating object proposals based on these fixations, and 3) tracking these objects to identify key moments of interaction changes. We build on the DEVA tracker~\cite{cheng2023tracking} for this implementation.

1. \textbf{Fixation Detection via In-Clip Consensus.} 
We detect fixations by analyzing eye gaze points across multiple frames. 
For each frame $t$, we generate object proposals $\{p_i^t\}$ using an image segmentation model (Segment Anything~\citep{kirillov2023segment}) 
given the gaze point as a prompt. An in-clip consensus 
is achieved by evaluating these proposals over a small temporal window of $n$ frames, $t, t+1, \ldots, t+n-1$. Fixations are identified as the object proposals that consistently appear in these frames, indicating sustained attention by the user.

Formally, following DEVA~\cite{}, let $Seg_{t+i}$ be the segmentation output at frame $t+i$, where $0 \leq i < n$. We define the object proposals at frame $t$ as the union of aligned segmentations across the temporal window:

\[
P_t = \bigcup_{i=0}^{n-1} \hat{S}_{t+i} = \{p_j | 0 < j \leq |P_t|\}
\]

The consensus output $C_t$ is then determined by selecting only those proposals $p_j$ that have a high overlap with proposals in subsequent frames:

\[
C_t = \{ p_j \in P_t | \text{IoU}(p_j, p_k) > \theta, 
\]
\[
\forall k \in \{t+1, \ldots, t+n-1\} \}
\]

where $\text{IoU}(p_j, p_k)$ is the Intersection-over-Union between the $j$-th proposal at frame $t$ and the $k$-th proposal in the subsequent frames, and $\theta$ is a threshold value. 


2. \textbf{Tracking fixated objects.} 
Using the fixation-based object proposals, we track objects across frames to monitor changes in user interaction. We use the DEVA tracker~\cite{}, which utilizes XMEM~\cite{} to propagate object masks from one frame to the next. Objects are continuously tracked as long as they remain within the field of view. An object stops being tracked if it leaves the field of view for more than $X$ frames.

As the video progresses, new objects may enter the field of view or become relevant due to changes in user focus. New object proposals are generated whenever a new fixation is detected outside the current consensus. These proposals are incorporated into the existing tracking framework using the same in-clip consensus method, ensuring the model dynamically adapts to the user's shifting focus.


3. \textbf{Temporal Boundary Detection.} We detect temporal boundaries by monitoring significant changes in the consensus object proposals over time. A temporal boundary is defined as the point in the video where there is a substantial change in the objects being tracked. Specifically, a boundary time $b$ is marked when the set of tracked objects at time $b$ differs from the set of tracked objects at the end of the previous segment by more than $Z\%$. This change indicates that the user's focus has shifted to different objects, signaling a new segment in the video.

\subsubsection{Speech-Based Temporal Segmentation}
We consider the use of the speech uttered by the user during the demonstration to temporally segment the video. We use the Whisper~\cite{} model to generate text segments, using the start and end times of the Whisper segments to define the temporal boundaries of segments in the video demonstration. An example speech segment is ``Make sure it's mixed up nicely, no chunks.''.

\subsection{Extracting key frames and textual knowledge.}\label{key_frame}


For each segment identified in Section~\ref{infer_intent}, we use the VLM to analyze the segment based on the user intent determined in Section~\ref{temp_seg}. This involves identifying key frames and generating descriptions of relevant information within each segment. We first provide the VLM with 30 subsampled frames from the segment. The VLM is then prompted to select the top-\(k\) most informative keyframes \( \{K_{i,1}, K_{i,2}, \ldots, K_{i,k}\} \) and generate a detailed caption \( C_i \) for the segment \( T_i \). For the gaze condition, we use visual prompting~\cite{} to highlight the gaze point on the image and reference this point in the prompt. For speech condition, we include any speech uttered by the user during the video segment. 


Note that our prompts are completely agnostic to the task or the user. We provide the prompt used to extract key frames and caption them in Listing~\ref{prompt_key_frame}. 


These outputs are stored in a database for in-context learning. 
The database stores each segment \( T_i \) as follows:

\[
\text{DB}(T_i) = \{(K_{i,1}, C_{ij}), (K_{i,2}, C_{ij}), \ldots, (K_{i,k}, C_{ij})\}
\]

Here, \(\text{DB}(T_i)\) denotes the database entries for segment \( T_i \), each consisting of a keyframe \( K_{i,j} \) and the corresponding caption \( C_{ij} \). We encode the keyframes for each segment into a visual vector, and the captions for each segment into a textual vector (CLIP~\citep{radford2021learning}) using a text encoder (OpenAI Embeddings Model~\cite{openaiembeddings}).
Given a new query, this structured storage facilitates efficient retrieval and use of keyframes and captions for retrieval-augmented generation.

\subsection{Inference on new questions}\label{inference}


Given a text query and corresponding visual input, such as an image and gaze data, we first generate a caption for the image using a captioning model. For offline captioning, we use GPT-4o, while Florence 2~\cite{xiao2023florence} is employed for real-time inference. 

The caption is encoded using a text encoder, and the image is processed through a visual encoder to obtain its visual embedding. We then compute the cosine similarity between the text embedding and image embedding with each segment's embeddings in the database. We select the top-\(k\) highest-scoring entries to use as context for retrieval-augmented generation.

The similarity score \(s\) for each entry is calculated as follows:

\[
s = \lambda_{\text{textual}} \cdot s_{\text{textual}} + \lambda_{\text{visual}} \cdot s_{\text{visual}}
\]

where:
- \(s_{\text{textual}}\) is the similarity score based on the text embedding,
- \(s_{\text{visual}}\) is the similarity score based on the visual embedding, and
- \(\lambda_{\text{textual}}\), and \(\lambda_{\text{visual}}\) are weighting factors that determine the contribution of each type of embedding to the overall similarity score.
We append the retrieved segment images and captions to the VLM prompt for additional context for question answering. 

Our prompt for inference on new questions is provided in Listing~\ref{prompt_inference}.

\subsection{Prompts}
We provide the prompts used in MICA in Listings 1-2.

\subsection{Key Frame Selection}\label{sec:key_frame_selection}

We provide examples of the key frames selected by GPT4o during the key frame selection and captioning step in MICA in Figures~\ref{fig:key_frame_selection_examples1}-~\ref{fig:key_frame_selection_examples4}.

\section{Data Collection}\label{supp_data_collection}
    
We conducted data collection using the HoloLens 2 device, where participants were recruited to perform user demonstrations and participate in evaluation episodes. We focused on three distinct task categories to assess user personalization from a single demonstration: organizing a room, shopping, and completing a morning routine. The data collection process was divided into two main phases. 

\begin{enumerate}
    \item \textbf{Demonstration Phase:} Participants were asked to perform one of three task types based on their personal preferences (Figure~\ref{fig:system_plot}; Left). 
    \item \textbf{Evaluation Phase:} Participants interacted with a live assistant that was provided with another participant's demonstration to generate ecologically valid questions (Figure~\ref{fig:system_plot}; Right).
\end{enumerate}

Data collected during the first phase served as the demonstration data, while the second phase data was used to evaluate the assistant's performance.  We recruited 10 participants (6 male, 4 female). Participants were compensated with a gift card. 

\subsection{Demonstration Phase}
\label{sec:demonstrationphase}
In the demonstration phase, participants were tasked with demonstrating one of three types of tasks. Each participant was given broad guidelines to follow but was allowed to complete the tasks according to their personal preferences. The tasks included:

\begin{itemize}
    \item \textbf{Organizing a Room:} Participants were asked to organize a set of objects on a shelving unit in one of three ways: (1) according to their personal preferences, (2) by object type, or (3) by color.
    \item \textbf{Shopping:} Participants were asked to shop for items based on one of the following scenarios: (1) their own needs or wants, (2) gathering materials to build a birdhouse, or (3) preparing for a toddler’s birthday party.
    \item \textbf{Morning Routine:} Participants were instructed to prepare specific juices, supplements such as protein powder and chia seeds (with varying scoop counts), and arrange specific types of supplements from a pill bottle.
\end{itemize}

These tasks—organizing a room, shopping, and executing a morning routine—reflect real-world scenarios that involve personal preferences and complex decision-making, making them ideal for studying the personalization of Vision-Language Models (VLMs). Each task varies significantly based on individual approaches, whether it’s arranging items, selecting products, or managing daily routines, capturing a broad range of user interactions. This variety provides rich data for understanding how VLMs can be tailored to meet the unique needs and preferences of users in everyday contexts.

It is important to note that even when a specific objective was given, such as organizing objects by type, participants had the freedom to group and place items as they saw fit. We instruct the participants to carry out the task naturally, and to ``speak aloud'' their intents as they demonstrated the task. We provide the full participant instructions for phase 1 in the appendix Listing~\ref{task_instruction_phase1}.

We recorded speech, RGB, eye gaze, and hand pose using the hl2ss~\citep{dibene2022hololens} software in Python. We transcribe all speech to text using Whisper~\citep{radford2022whisper}.

\begin{figure}
    \centering
    \includegraphics[width=0.6\columnwidth]{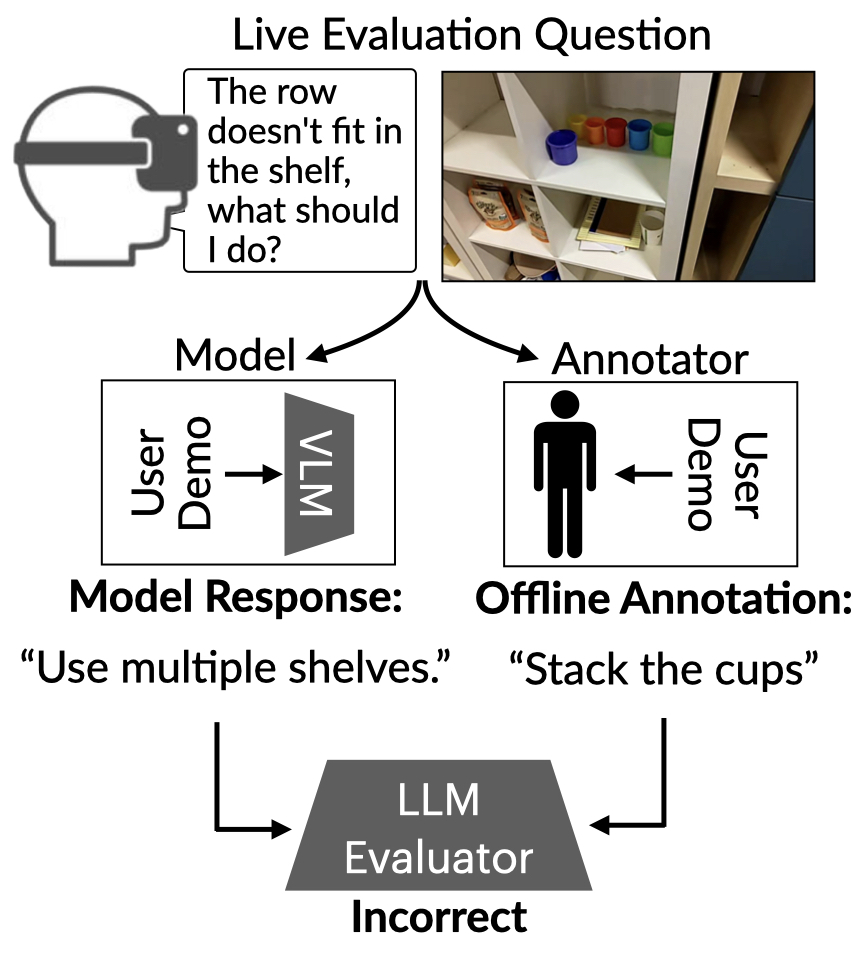}
    \caption{We conducted an offline evaluation using real questions asked during the real-time evaluation phase. Each question was paired with the corresponding user demonstration provided to the live assistant during the evaluation. Ground truth answers were annotated offline based on these demonstrations. To assess model performance, we queried models with the same demonstrations and compared their responses to the human-provided answers using a LLM as an evaluator.}
    \label{fig:offline_eval}
\end{figure}



\subsection{Live Evaluation Phase}
\label{sec:liveevalphase}
To generate evaluation data, we aim to: 1) \textbf{create ecologically-valid queries} that a user might ask while interacting with an AI assistant to complete a task, and 2) obtain \textbf{user-specific queries and responses} based on a particular user demonstration. 
In the evaluation phase, we have a live assistant guide a new user to complete a task, based on a demonstration collected by a different user. Each participant was given a broad overview of the task (e.g., ``Your task is to interact with an AI assistant to determine how to organize the items in the room''). The participant is able to interact with the assistant by asking it spoken questions, with the goal of replicating what the other user did during the demonstration. We provide the full participant instructions for phase 2 in the appendix Listing~\ref{task_instruction_phase2}.

\textbf{Real-time interactive system.} 
We present a real-time interactive system that answers user questions based on demonstrations provided through sensor inputs from the HoloLens 2 device. Our system leverages hl2ss~\citep{dibene2022hololens} to stream through TCP ports and synchronize these sensor inputs, interfacing them with our models in Python. Users initiate a question by saying ``Hey agent,'' after which they have a predefined period to ask their question while the user's current egocentric RGB view is recorded. The audio input is then transcribed to text using Whisper, and used to query our assistant.

The model retrieves relevant context from a pre-defined database initialized from a user demonstration, processed using our gaze and speech method as described in the main paper. The question and user image is encoded, per Section~\ref{inference}. This context is utilized for retrieval-augmented generation using GPT4o to formulate an answer, which is subsequently converted to speech with OpenAI's TTS and delivered to the user through a speaker. We provide GPT4o with a chat history of questions and user images throughout the interaction.


\subsubsection{Offline evaluation.} 
\label{sec:offlineeval}
To benchmark the different models in their ability to extract information from user demonstrations and answer ecologically valid questions, we developed an offline evaluation based on real questions asked during the real-time evaluation phase. For each question, we paired it with the corresponding user demonstration that was provided to the live assistant during the evaluation episode. This approach allows us to assess how well each model can interpret and respond to questions grounded in actual user interactions, thereby providing a realistic measure of their performance in understanding and utilizing user input in real-world scenarios.

\subsubsection{LLM-Match: Evaluating Answer Correctness}\label{sec:llmmatch}
Evaluating open-vocabulary answers in QA is challenging due to the variety of possible correct responses. While human evaluation is a reliable method, it is often too slow and costly. To address this, we use an LLM to assess the correctness of answers generated by agents. We adapt the evaluation method from OpenEQA~\cite{OpenEQA2023} for this purpose. Given a question $Q_i$, a human-provided reference answer $A^*_i$, and the model-generated answer $A_i$, the LLM assigns a score $\sigma_i \in \{1, 2, 3\}$. In this scoring system, a score of 1 corresponds to an incorrect response, 3 denotes a correct response, and 2 indicates a partially correct response. We compute the overall LLM-based correctness metric (LLM-Match) as follows:

\begin{equation}
C = \frac{1}{N} \sum_{i=1}^{N} \frac{\sigma_i - 1}{2} \times 100\%.
\end{equation}

\textbf{Dataset statistics.} In total, our evaluation set includes 10 participants, 32 user demonstrations, and 415 evaluation questions. 

\textbf{Human annotations.} We annotate each of the questions offline based on the user demonstration to get the ground truth answer $A^*$. The annotator watches the user demonstration and then annotates the ground truth answer based on the demonstration. Some questions may have multiple answers to be correct. We flag and remove from the evaluation set highly ambiguous questions. 

\subsection{Task Instructions}
We provide task instruction given to participants in Listings 3-4.


\section{Additional Results}\label{supp_add_results}

\subsection{Results in Table Format}
In Table~\ref{tab:model_comparison}, we provide the results from the main paper in table format.

\begin{table*}[ht]
\centering
\small
\begin{tabular}{lccccc}
\toprule
\textbf{Condition} & \textbf{GPT4o} & \textbf{GPT4o Mini} & \textbf{VILA-3b} & \textbf{LLaVa-OV-7b} & \textbf{VILA-3b} \\
\textbf{} & \textbf{} & \textbf{} & \textbf{w/ GPT4o Ctx} & \textbf{w/ GPT4o Ctx} & \textbf{} \\
\midrule
Zero Shot & 28.92 (±2.10) & 19.28 (±1.82) & 18.61 (±1.85) & 14.14 (±1.66) & 18.80 (±1.83) \\
Clip Clustering & 49.28 (±2.31) & 36.14 (±2.17) & 32.38 (±2.16) & 26.67 (±2.10) & -- \\
Frames-As-Context & 48.43 (±2.28) & 40.72 (±2.22) & 28.04 (±2.07) & 22.46 (±2.01) & -- \\
Eye Gaze w/ Inf. Intent & 52.41 (±2.30) & -- & -- & -- & -- \\
Eye Gaze & 55.37 (±2.49) & 40.96 (±2.23) & 37.97 (±2.27) & 30.65 (±2.17) & 18.07 (±1.75) \\
Speech & 59.40 (±2.23) & 51.57 (±2.22) & 34.86 (±2.20) & 29.53 (±2.16) & 20.36 (±1.86) \\
Eye Gaze + Speech & 62.16 (±2.30) & 51.08 (±2.26) & 40.57 (±2.29) & 34.00 (±2.25) & 21.57 (±1.90) \\
Eye Gaze + Speech + Summary & 66.51 (±2.15) & -- & -- & -- & -- \\
\bottomrule
\end{tabular}
\caption{Performance across different input conditions for each model. Values are mean (± standard error).}
\label{tab:condition_comparison}
\end{table*}

\subsection{Model accuracy broken down by task type}
\begin{figure}[htbp]
    \centering
    \includegraphics[width=0.8\columnwidth]{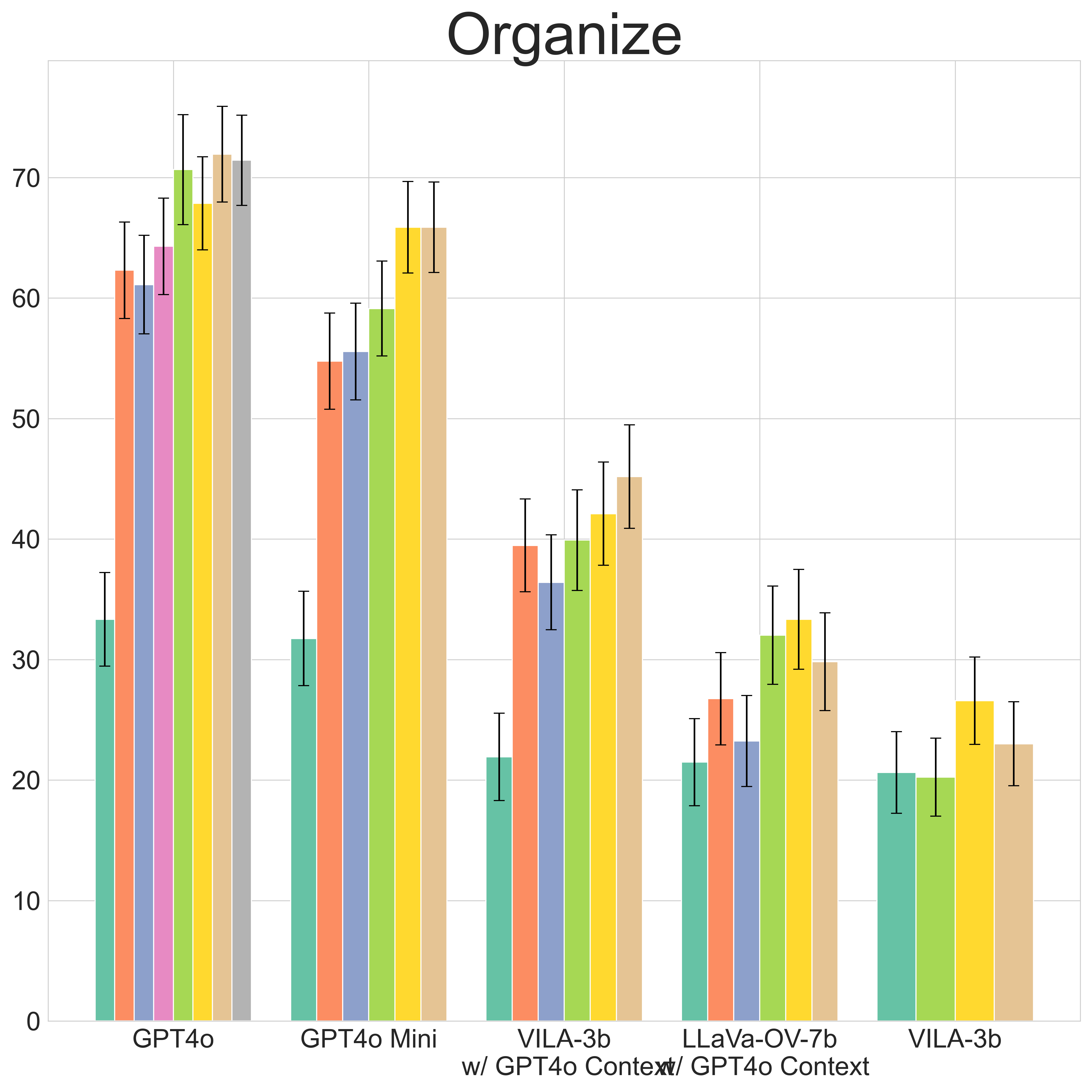}
    
    \vspace{5mm} 
    
    \includegraphics[width=0.8\columnwidth]{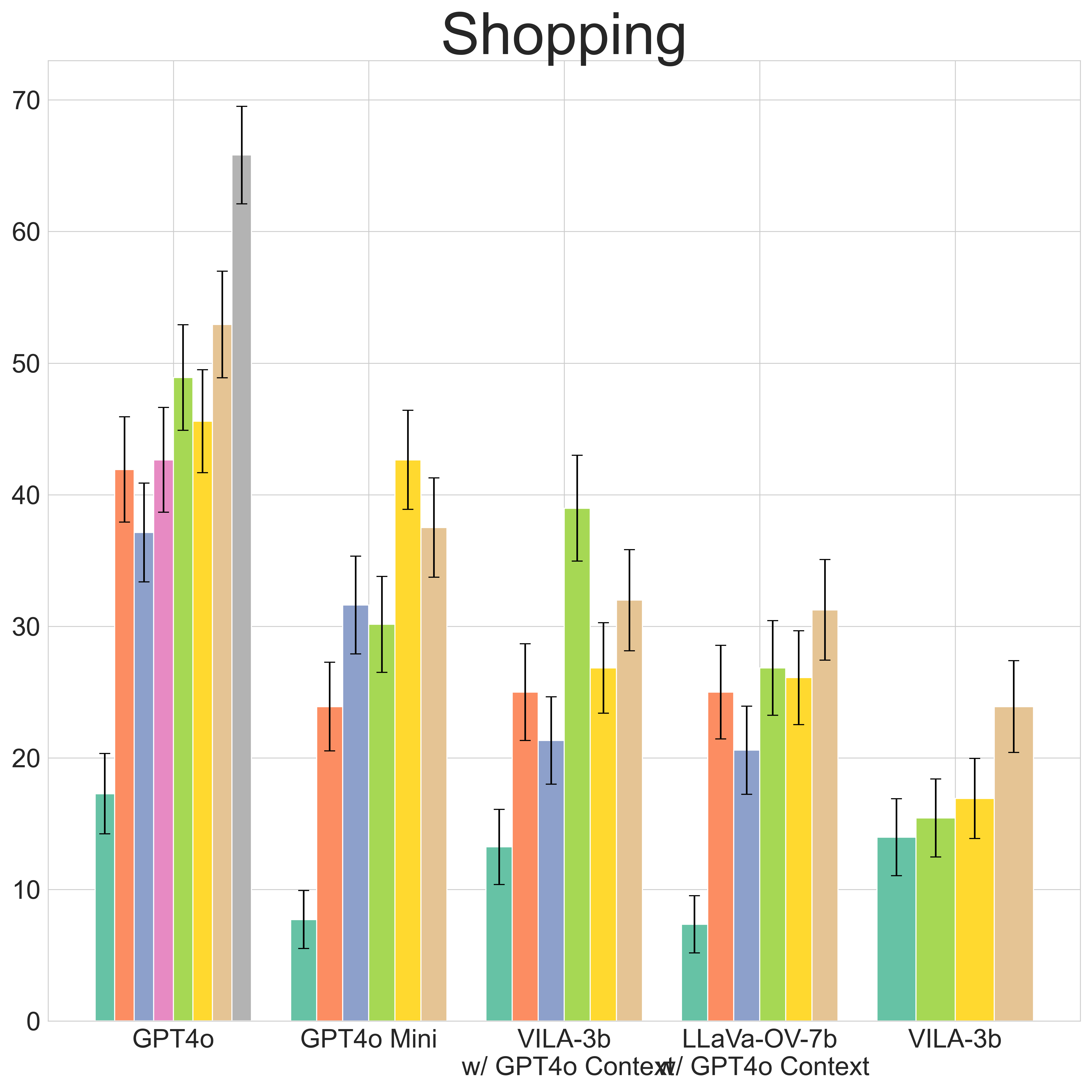}
    
    \vspace{5mm} 
    
    \includegraphics[width=0.8\columnwidth]{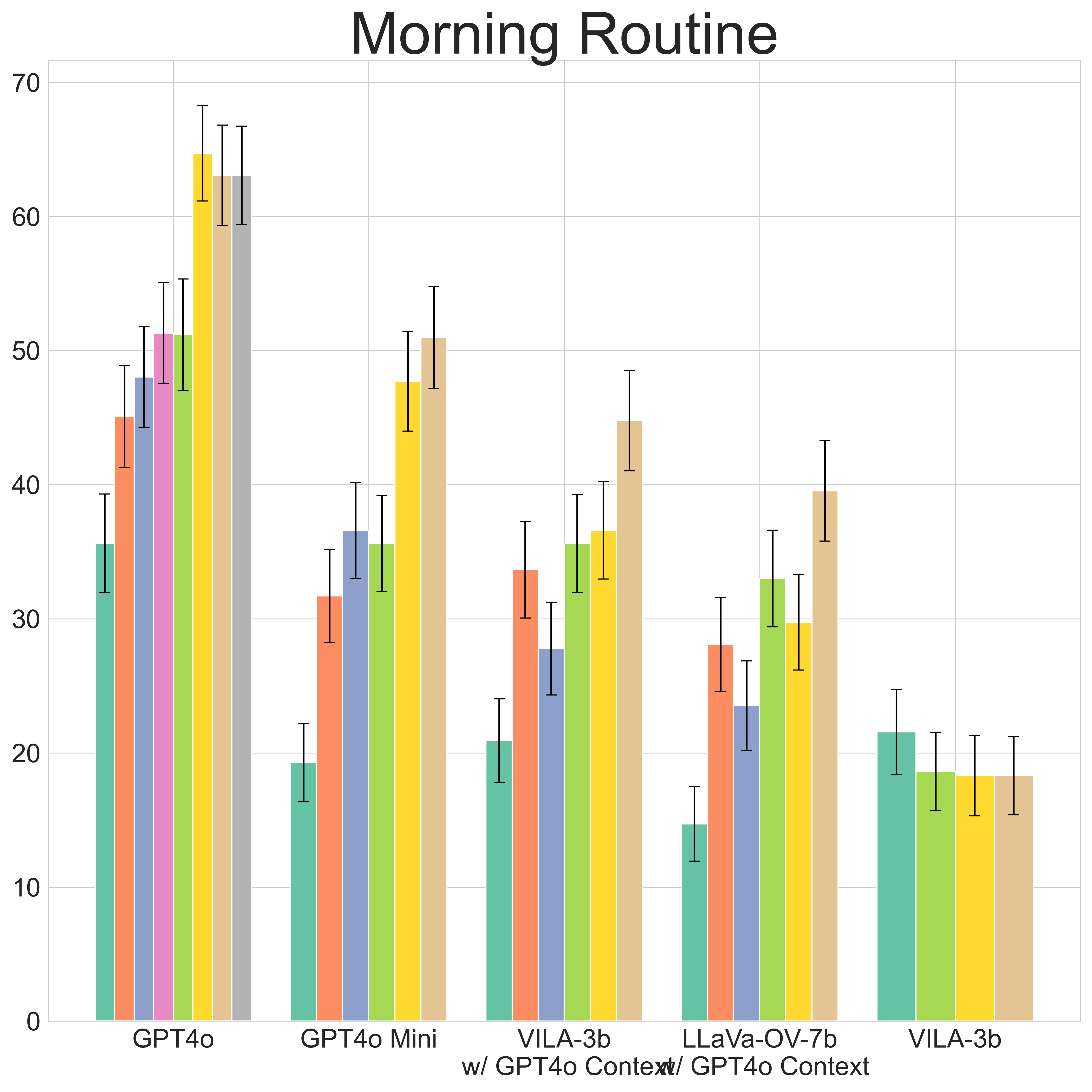}

    \caption{
    Performance of base models (GPT4o, GPT4o-mini, VILA, LLaVA) across different context extraction methods (Zero Shot, CLIP Clustering, Frames as Context, Eye Gaze, Speech, Eye Gaze + Speech) on the annotated evaluation set. 
    Results are broken down by task: (a) Organize, (b) Shopping, and (c) Morning Routine. 
    We report mean $\pm$ standard error across evaluation questions (n=413).
    }
    \label{fig:taskwise_performance}
\end{figure}

In Figure~\ref{fig:taskwise_performance}, we show model performance in the evaluation dataset broken down by task type.

\subsection{Which combinations of input modalities and feature encoders provide the best performance for retrieval-augmented generation?}

In Table~\ref{tab:ablations}, we compare various visual encoders and input types for retrieving context from the database to supply to GPT4o for retrieval-augmented generation. The highest accuracy was achieved by the combined approach of CLIP and OpenAI Embeddings (56.6{\scriptsize $\pm$ 2.4}), which equally weights visual and textual inputs (\(\lambda_{\text{visual}} = 0.5\), \(\lambda_{\text{textual}} = 0.5\)). Comparisons between unimodal configurations (e.g., CLIP image-only vs. OpenAI Embed text-only, U test, \(p\text{-val}=0.6269\)) suggest that \textbf{neither visual nor textual inputs alone dominate in performance, underscoring the balanced contribution of both visual and textual cues in ecological QA tasks.} 

When comparing CLIP + OpenAI Embed with AlphaCLIP + Eye Gaze, the results show no significant difference (\(t\text{-stat}=90109.5000\), \(p\text{-val}=0.2030\)). This suggests that the use of eye gaze in AlphaCLIP visual encoding does not substantially improve retrieval performance in this setting. Despite eye gaze's potential for contextual relevance, its integration here did not yield significant improvements, possibly due to the text content already incorporating eye gaze cue information. Similarly, ImageBind, which jointly encodes images, text, audio, depth, thermal, and IMU data, did not perform as well as CLIP + OpenAI Embed, emphasizing that the inclusion of more sensory inputs does not necessarily translate into better performance. The joint encoding strategy of ImageBind appears less effective, potentially due to the complexity and noise introduced by integrating heterogeneous data types that do not directly contribute to the specific QA task. These findings highlight the critical role of strong and well-aligned visual and textual encodings, as achieved by CLIP + OpenAI Embed, in improving the retrieval of relevant context in ecological QA settings.

\begin{table}[h!]
\small
\setlength{\tabcolsep}{2pt} 
\setlength{\tabcolsep}{2pt} 
\caption{Comparison of multimodal and unimodal configurations on ecological QA tasks. The highest accuracy is achieved by CLIP + OpenAI Embeddings with balanced visual and textual inputs (\(\lambda_{\text{visual}} = 0.5\), \(\lambda_{\text{textual}} = 0.5\)). Eye gaze integration in AlphaCLIP does not significantly improve performance, and ImageBind's multimodal approach shows lower accuracy, highlighting the effectiveness of strong, aligned visual and textual encodings. We use the gaze only condition for all ablations.}\label{tab:ablations}
\begin{tabular}{@{}lccc@{}}
 \\ & \multicolumn{1}{c}{Accuracy} & \multicolumn{1}{c}{$\lambda_{\text{visual}}$} & \multicolumn{1}{c}{$\lambda_{\text{textual}}$} \\
\addlinespace[0.15cm] 
\hline
\addlinespace[0.15cm] 
 CLIP~\cite{radford2021learning} & \textbf{56.6}{\scriptsize $\pm$ 2.4} & 0.5 & 0.5 \\
 \hspace{0.5mm}+ OpenAI Embed~\cite{openaiembeddings} & & & \\
 CLIP~\cite{radford2021learning} & 51.9{\scriptsize $\pm$ 2.3} & 1.0 & 0.0 \\
 OpenAI Embed~\cite{openaiembeddings} & 53.5{\scriptsize $\pm$ 2.3} & 0.0 & 1.0 \\
 AlphaCLIP~\cite{sun2024alpha} & 52.5{\scriptsize $\pm$ 2.3} & 1.0 & 0.0 \\
 \hspace{0.5mm} + Eye Gaze & & & \\
 ImageBind~\cite{girdhar2023imagebind} & 48.7{\scriptsize $\pm$ 2.3} & 0.5 & 0.5 \\
\addlinespace[0.15cm]
 \hline
 \addlinespace[0.15cm] 
\end{tabular}
\end{table}

\begin{table}[h!]
\setlength{\tabcolsep}{2pt} 
\setlength{\tabcolsep}{2pt} 
\caption{Ablation study on the importance of modeling chat-image history in ecological QA tasks. Including chat-image history in the Eye Gaze + Speech condition improves accuracy, though the difference is not statistically significant (\(p = 0.2499\)). Incorporating question-answer pairs generated by GPT4o shows minor, nonsignificant improvements in accuracy.}\label{tab:QA_history}
\begin{tabular}{@{}lcc@{}}
 \\
 & \multicolumn{1}{c}{Accuracy} \\
\addlinespace[0.15cm] 
\hline
\addlinespace[0.15cm] 
 \textbf{Eye Gaze + Speech} & 62.2{\scriptsize $\pm$ 2.3} \\
 \hspace{0.5mm} w/ QA & 64.2{\scriptsize $\pm$ 2.2} \\
 \hspace{0.5mm} w/o chat-image history & 58.4{\scriptsize $\pm$ 2.3} \\
 \textbf{Eye Gaze} & 56.6{\scriptsize $\pm$ 2.4} \\
 \hspace{0.5mm} w/ QA & 58.2{\scriptsize $\pm$ 2.3} \\
\addlinespace[0.15cm]
 \hline
 \addlinespace[0.15cm] 
\end{tabular}
\end{table}

\subsection{How important is short term history in ecologically evaluation?}
User questions often depend on previous interactions (i.e., the chat history) to accurately contextualize the answer. In this section, we ask, `how important is modeling chat history in an ecological setting?'. We ablate the image-chat history in Table~\ref{tab:QA_history}. We observe a numerical decrease in accuracy when removing the chat-image history from our model (Eye Gaze + Speech condition; 62.2{\scriptsize $\pm$ 2.3} vs. 58.4{\scriptsize $\pm$ 2.3}), though this difference is not statistically significant (U test $p=0.2499$). These results suggest that while chat history may contribute to performance, its impact in this setting is not definitive, highlighting the need for further investigation into the conditions under which history modeling is most beneficial.

\subsection{Can VLMs ask good questions to get additional context from the user?}\label{sec:QA_results}
We additionally test the ability of GPT4o to generate questions to get additional clarifying information from the user about their preferences.
We find that appending these question-answer pairs to the context of GPT4o provides minor, nonsignificant improvements, as shown in Table~\ref{tab:QA_history}. This indicates that using GPT4o to generate clarifying information from the user does not provide significant advantages over running the demonstration only, both for the gaze and speech condition.

\subsection{Retrieval Accuracy}
\begin{figure}
    \centering
    \includegraphics[width=1.0\columnwidth]{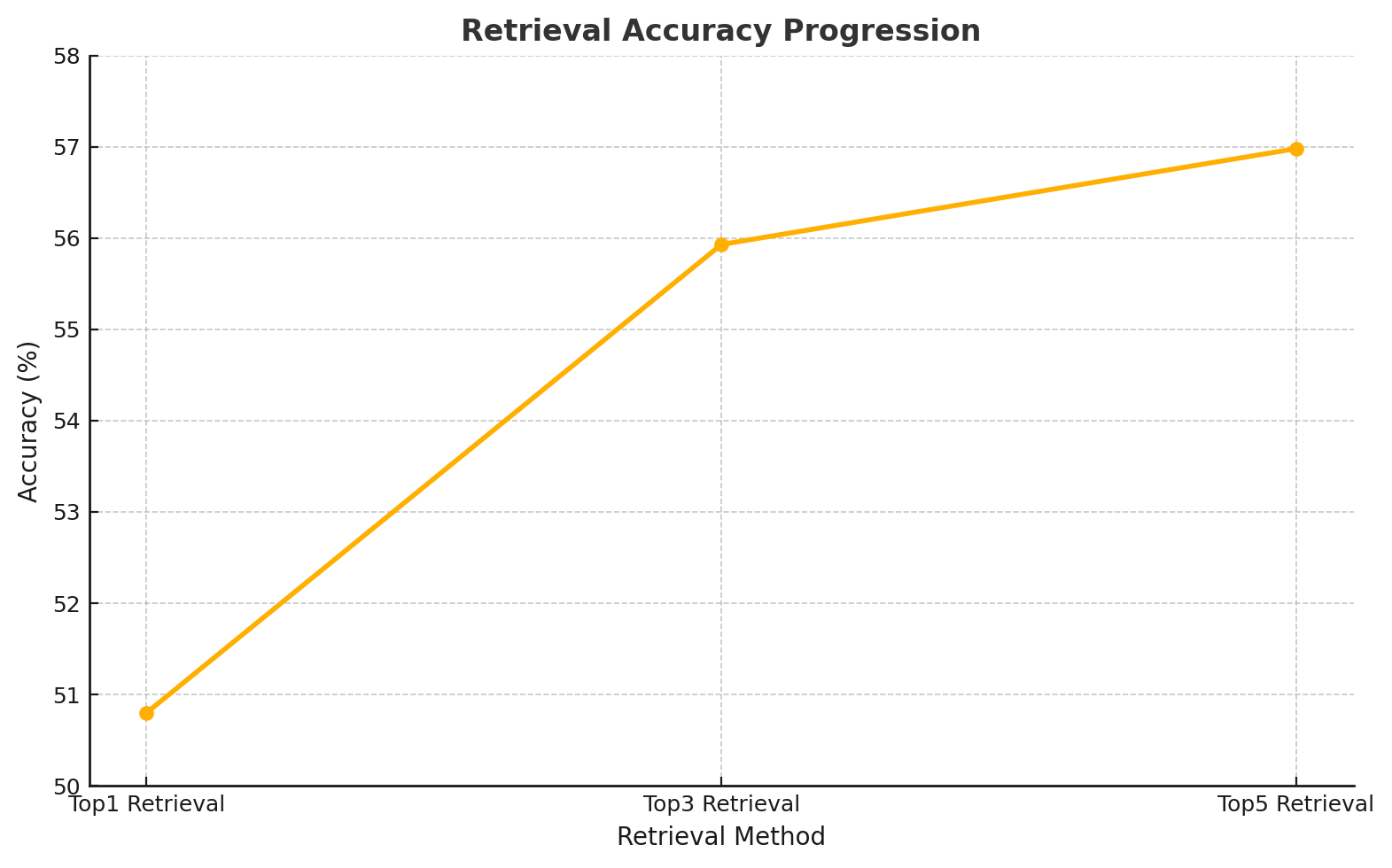}
    \caption{Retrieval accuracy progression when retrieval the top-k segments for context. The solid line illustrates the improvement from Top-1 to Top-5 example retrieval.}\label{fig:retrieval_accuracy}
\end{figure}

\begin{figure*}
    \centering   \includegraphics[width=0.9\textwidth]{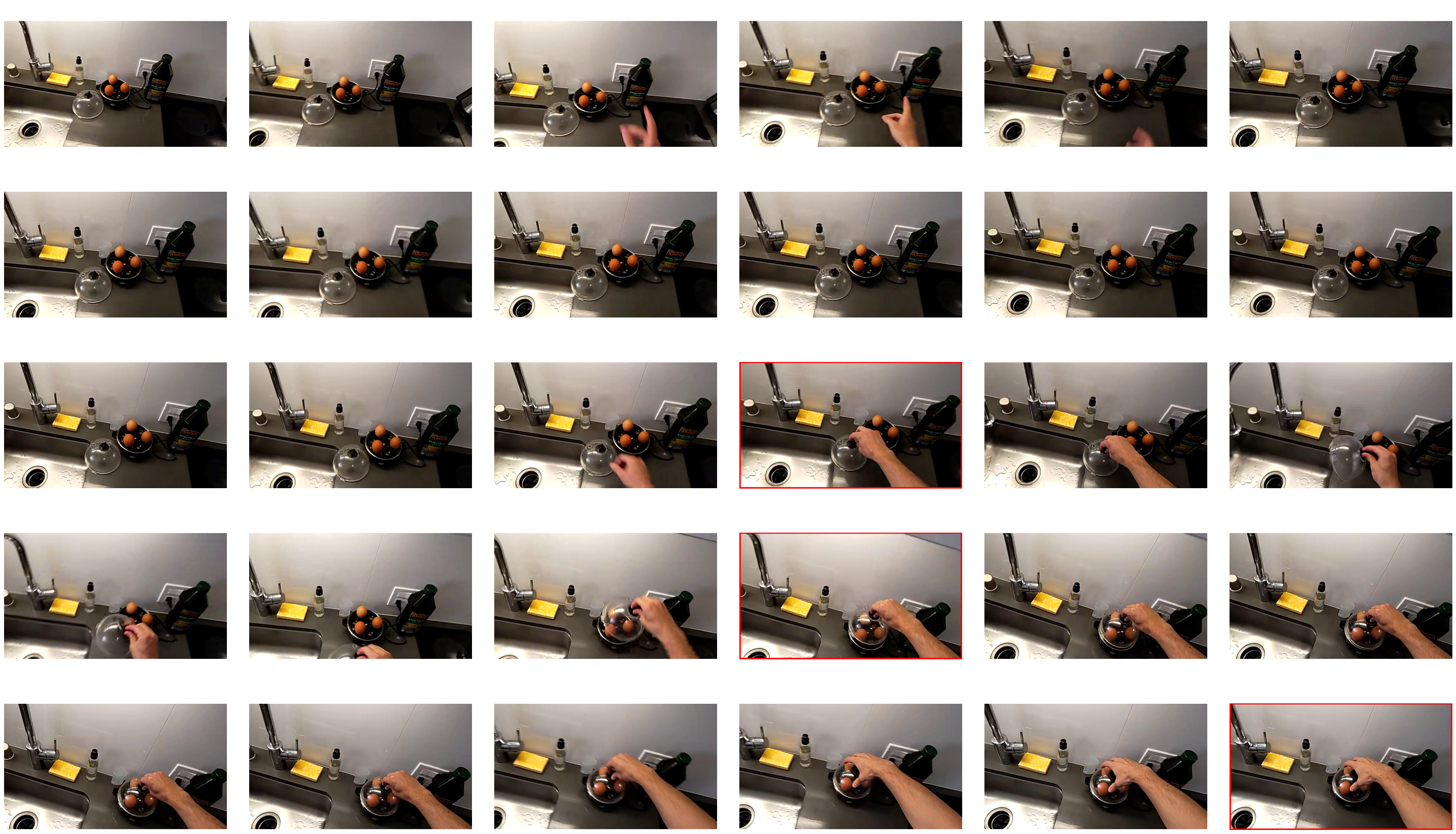}
    \caption{\textbf{Examples of key frame selection from the making breakfast video.} We provide the VLM with 30 unsampled frames, asking it to select the top-k frames. Red highlighted frames indicate selected key frames.}
\label{fig:key_frame_selection_examples1}.
\end{figure*}

\begin{figure*}
    \centering   \includegraphics[width=0.9\textwidth]{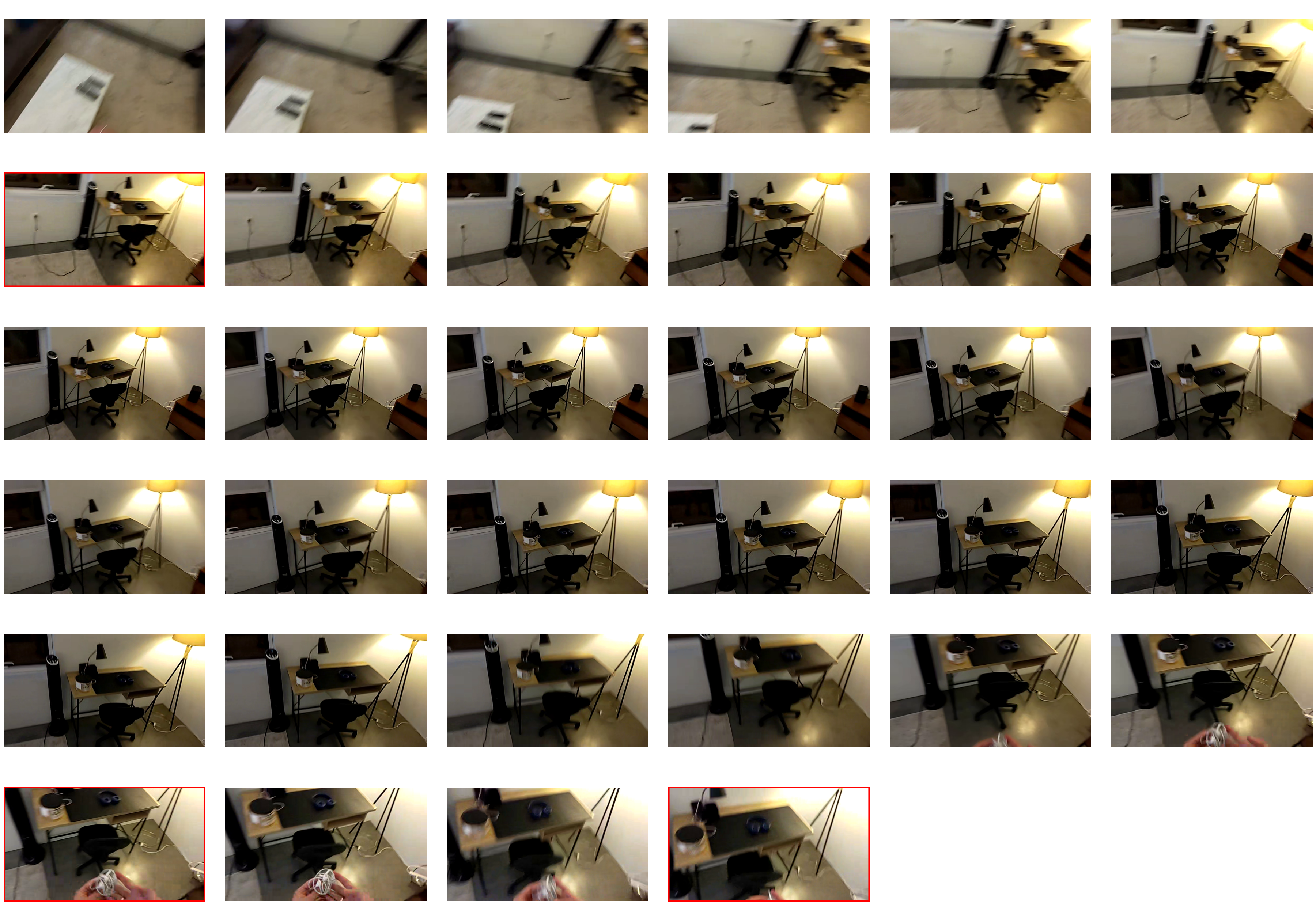}
    \caption{\textbf{Examples of key frame selection from the organizing apartment video.} We provide the VLM with 30 unsampled frames, asking it to select the top-k frames. Red highlighted frames indicate selected key frames.}
\label{fig:key_frame_selection_examples2}.
\end{figure*}

\begin{figure*}
    \centering   \includegraphics[width=0.9\textwidth]{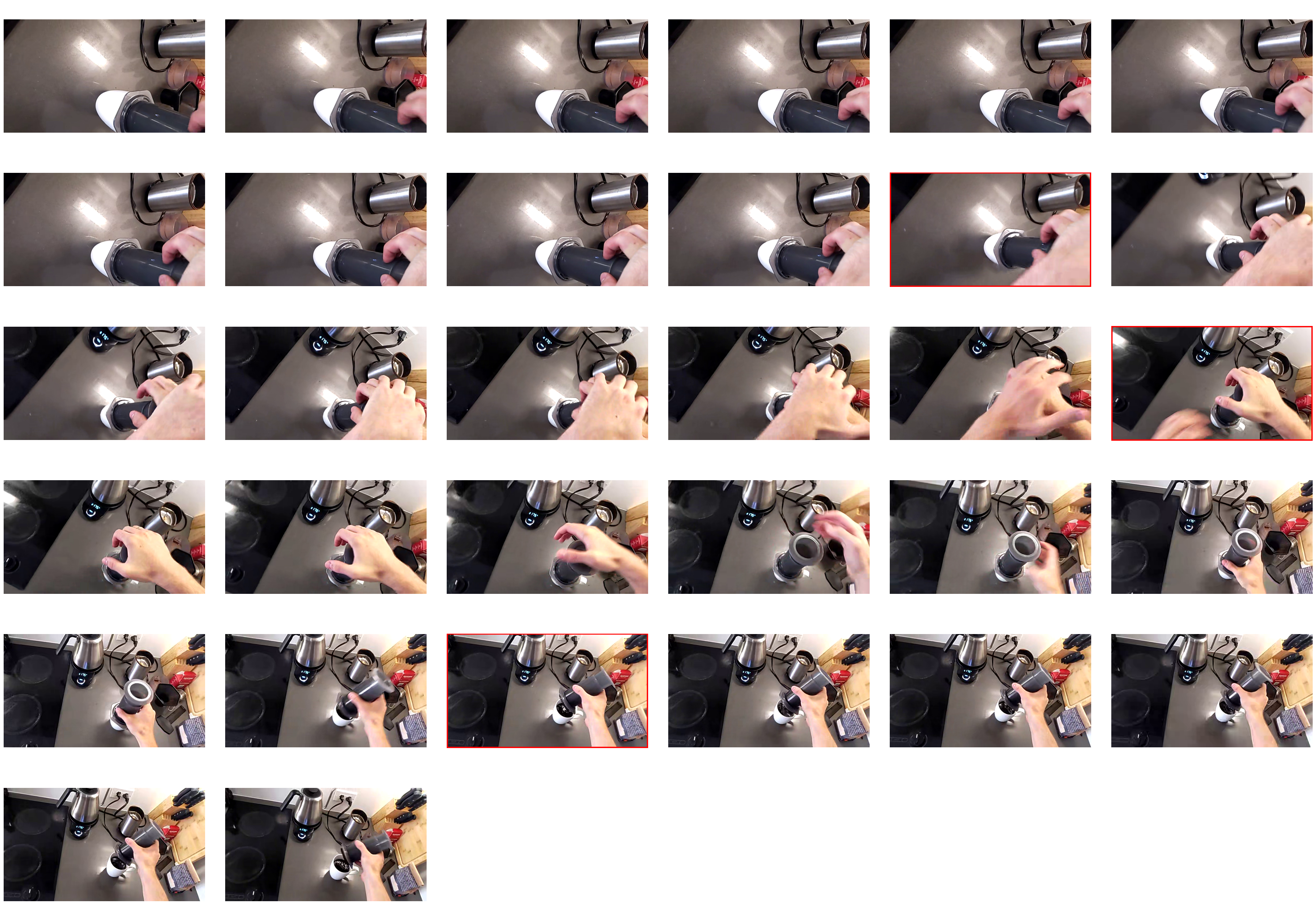}
    \caption{\textbf{Example of key frame selection from the making breakfast apartment video.} We provide the VLM with 30 unsampled frames, asking it to select the top-k frames. Red highlighted frames indicate selected key frames.}
\label{fig:key_frame_selection_examples3}.
\end{figure*}

\begin{figure*}
    \centering   \includegraphics[width=0.9\textwidth]{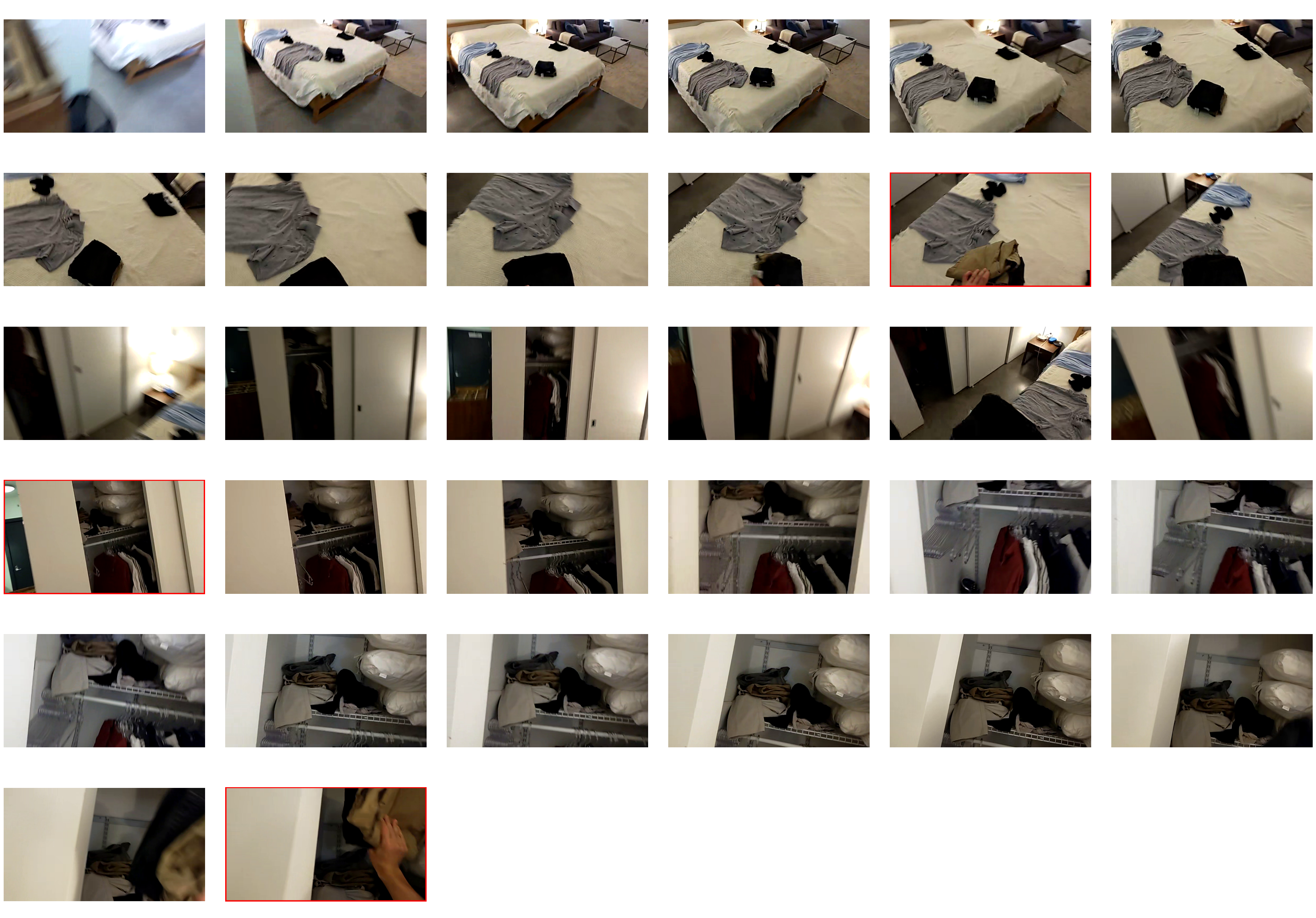}
    \caption{\textbf{Examples of key frame selection from the organizing apartment video.} We provide the VLM with 30 unsampled frames, asking it to select the top-k frames. Red highlighted frames indicate selected key frames.}
\label{fig:key_frame_selection_examples4}.
\end{figure*}

We provide an analysis of retrieval accuracy in Figure~\ref{fig:retrieval_accuracy}. We find that performance plateaus at retrieving the top 3 segments. 

\section{Behavioral Coding}\label{sec:utterance_categories}

\subsection{Demonstration Utterance Coding}

We code the utterances into the following categories (examples come directly from our dataset):
\begin{enumerate}[label=\textbf{\arabic*.},leftmargin=*]

\item \textbf{Purpose of the Task:} This category includes statements that explain the objective or goal of the task. These utterances typically provide a high-level overview of what the speaker intends to demonstrate or accomplish. For example:
    \begin{itemize} 
        \item ``Okay, I'm going to demonstrate my morning routine.''
        \item ``Okay, today I'm going to show you how to organize the items on these tables based on my preferences.''
    \end{itemize}

    \item \textbf{Instructions to the Learner:} These are directives given to the learner to focus their attention or to observe the actions being performed. They are aimed at guiding the learner on how to follow along with the task. Examples include:
    \begin{itemize}
        \item ``Take note of what I'm doing.''
        \item ``Pay close attention.''
    \end{itemize}

    \item \textbf{Action Commands:} This category contains direct instructions that tell the learner or listener what specific action to take. These commands are usually clear and concise. For instance:
    \begin{itemize}
        \item ``Take one of these, put it in the water.''
        \item ``Open it up.''
        \item ``Stir the water.''
    \end{itemize}

    \item \textbf{Sequential Steps:} These utterances describe the order in which actions should be performed. They help in outlining the sequence of tasks that need to be completed. Examples include:
    \begin{itemize}
        \item ``First, I'm going to stack all these cups up.''
        \item ``Next, I'm going to pick up the small play basketball model here.''
        \item ``Finally, take an empty cup, like this one.''
    \end{itemize}

    \item \textbf{Clarifying Actions:} This category includes statements that provide further explanation or details about a given action, often clarifying what needs to be done. Examples are:
    \begin{itemize}
        \item ``So we're going to take two of these, two of these ones, put them on the plate.''
        \item ``Okay, and then two of these, one and two, in the water, okay, close it, and now we're...''
    \end{itemize}

    \item \textbf{Rationale:} These utterances explain the reason behind a specific step or action, providing context or justification for why it is necessary. For example:
    \begin{itemize}
        \item ``Make sure it's mixed up nicely, no chunks.''
        \item ``Because I don't have enough hands, so I'm putting this strawberry inside the cup.''
    \end{itemize}

    \item \textbf{Task Completion:} This category indicates the conclusion of a task or series of actions, signaling that no further steps are required. Examples include:
    \begin{itemize}
        \item ``And we're done.''
        \item ``Task complete.''
    \end{itemize}

    \item \textbf{Interim Confirmation:} These utterances provide confirmation of intermediate steps or actions, often used to acknowledge progress or correctness. For instance:
    \begin{itemize}
        \item ``Okay, great.''
        \item ``Okay, that's mixed up.''
    \end{itemize}

    \item \textbf{Self-Questions:} This category includes questions the speaker asks themselves during the task, often reflecting uncertainty or prompting themselves to think. Examples are:
    \begin{itemize}
        \item ``Shall I start?''
        \item ``What else do I need?''
    \end{itemize}

    \item \textbf{Responses:} These utterances are answers to questions or confirmations of actions, often used to acknowledge understanding or correctness. Examples include:
    \begin{itemize}
        \item ``Yep, it works.''
        \item ``No, that's a toy.''
    \end{itemize}

    \item \textbf{Observational Comments:} This category consists of comments made during the task that note observations or provide additional thoughts about objects or actions. For example:
    \begin{itemize}
        \item ``This seems handy, so let me put this in the basket.''
        \item ``This looks like a glass cleaner probably not more very important right now.''
    \end{itemize}

    \item \textbf{Personal Preferences:} These utterances reflect the speaker's personal choices or preferences, often indicating likes or dislikes. Examples are:
    \begin{itemize}
        \item ``Not a big fan of decoration, so no.''
        \item ``I love snacks, so I want different kinds of snacks.''
    \end{itemize}

    \item \textbf{Error Identification:} This category includes statements that recognize mistakes or issues encountered during the task. For instance:
    \begin{itemize}
        \item ``Okay, this wasn't correct I guess.''
        \item ``Oh, I forgot to stack the yellow cup in this bigger stack.''
    \end{itemize}

    \item \textbf{Adjustments:} These utterances describe corrections or modifications made to rectify mistakes or to improve the task's execution. Examples include:
    \begin{itemize}
        \item ``So I have to redo it again.''
        \item ``I will put a bit of it in the cup.''
    \end{itemize}

    \item \textbf{Filler Words and Phrases:} This category includes non-essential words or phrases used during the task, often to fill gaps in speech or to signal thinking. Examples are:
    \begin{itemize}
        \item ``Okay.''
        \item ``Alright.''
    \end{itemize}

    \item \textbf{Repetitions:} These utterances involve repeating words or phrases for emphasis or to ensure understanding. Examples include:
    \begin{itemize}
        \item ``Birdhouse. Birdhouse.''
        \item ``Okay, okay.''
    \end{itemize}

\end{enumerate}

\subsection{Question Coding}\label{sec:question_categories}

We code the questions into the following categories (examples come directly from our dataset):

\begin{enumerate}[label=\textbf{\arabic*.}, leftmargin=*]

    \item \textbf{Clarification/Confirmation Questions:} These questions seek confirmation or clarification about a previous action or instruction. They are typically used to ensure correctness or understanding. Examples include:
    \begin{itemize}
        \item ``Is this correct?''
        \item ``Did I take any of these pills last time?''
        \item ``Is this the correct one?''
        \item ``Am I done?''
        \item ``Did I put chia seeds into the drink?''
    \end{itemize}

    \item \textbf{Instructional/Procedural Questions:} These questions request instructions or guidance on the next steps or actions to take. They help in understanding the procedure or sequence of actions. Examples are:
    \begin{itemize}
        \item ``What should I do next?''
        \item ``What should I do with these?''
        \item ``How should I mix it?''
        \item ``What should be the next step?''
    \end{itemize}

    \item \textbf{Quantitative Questions:} These questions focus on the amount, number, or measurement of items, often to determine the quantity required. Examples include:
    \begin{itemize}
        \item ``How many scoops should I put for chia seeds?''
        \item ``How much of this protein powder should I take?''
        \item ``How many of the protein powder, how much of it should I use?''
        \item ``How many cups should I get?''
    \end{itemize}

    \item \textbf{Choice/Selection Questions:} These questions involve making a choice or selecting between multiple options. They are used to determine a preference or decision. Examples are:
    \begin{itemize}
        \item ``Which cup should I put the protein powder in?''
        \item ``Which shelf should I put the ceramic mugs onto?''
        \item ``Which pill should I take today?''
        \item ``Which item should I get started?''
    \end{itemize}

    \item \textbf{Location/Identification Questions:} These questions seek to identify or locate an item or place. They help in finding or confirming the position of objects. Examples include:
    \begin{itemize}
        \item ``Where should I put the pills?''
        \item ``Where is the pan?''
        \item ``Is this the shelf you are referring to?''
        \item ``Where should I keep these fruits?''
    \end{itemize}

    \item \textbf{Recall/Memory Questions:} These questions ask the AI to remember or recall past actions or information, often to verify if something has already been done. Examples are:
    \begin{itemize}
        \item ``Do you remember what powder I put for this drink?''
        \item ``Did I buy this item already?''
        \item ``Did I take the vitamins last time?''
    \end{itemize}

    \item \textbf{Verification Questions:} These questions verify the accuracy or correctness of a task or item, ensuring that actions have been performed correctly. Examples include:
    \begin{itemize}
        \item ``Is this the right hammer?''
        \item ``Does this look neat for you?''
        \item ``Is this placement correct?''
        \item ``Is it the right one?''
    \end{itemize}

    \item \textbf{Descriptive/Inquiry Questions:} These questions seek descriptions or inquire about specific characteristics of objects or actions. They help in understanding details or functions. Examples are:
    \begin{itemize}
        \item ``Describe that drug for me.''
        \item ``What are these pills for?''
        \item ``What do I do with the protein powder, vitamins, and the flaxseed?''
        \item ``What should I take today?''
    \end{itemize}

    \item \textbf{Opinion/Preference Questions:} These questions seek the AI's opinion or recommendation, often asking for subjective feedback or suggestions. Examples include:
    \begin{itemize}
        \item ``Does this look good for my morning routine?''
        \item ``Should I buy this drawing pen?''
        \item ``What would be the best way to organize this stuff?''
    \end{itemize}

    \item \textbf{Miscellaneous Questions:} These include various questions that don’t fit neatly into the above categories. They can be open-ended or cover a range of topics. Examples are:
    \begin{itemize}
        \item ``Can you speak English?''
        \item ``What's up?''
        \item ``Shall I take the whole or just a few pieces of wipes?''
        \item ``What's going on?''
    \end{itemize}

    \item \textbf{Behavioral Observation Questions:} These questions ask the AI to confirm its observational capabilities, often checking if the AI has noticed specific actions. Example:
    \begin{itemize}
        \item ``Did you see that I pour cranberry juice for the drink?''
    \end{itemize}

    \item \textbf{Non-Task-Related Commentary:} This isn’t a direct request or question but rather a commentary that might invite the AI to respond empathetically or adjust its recommendations. Example:
    \begin{itemize}
        \item ``This sounds terrible.''
    \end{itemize}

    \item \textbf{Self-Directed Questions:} This question seems to be more of a self-reflection than a direct question to the AI, but it’s posed in a way that invites the AI to confirm, showing a blend of internal dialogue and external communication. Example:
    \begin{itemize}
        \item ``I think I already done this, right?''
    \end{itemize}

\end{enumerate}





\onecolumn\begin{lstlisting}[caption={Prompt template for key frame extraction and captioning},captionpos=t,label={prompt_key_frame}] 
You are an AI assistant tasked with analyzing video frames to identify key moments that are most descriptive of the overall task being performed. The inputs to this task include a sequence of video frames and overall intent. Your goal is to provide a detailed description of the task segment, identify the top three key frames that are most informative for understanding that segment, and a detailed description of each key frame.

### Inputs:
1. **Segment History**: A description of each segment that came before the current video segment you are analyzing.
2. **Video Frames**: A sequence of images representing different frames from a video segment. (IF IMPLICIT CONDITION) The user's gaze location during the query is annotated on the image with a purple circle. The user's right and left hand locations are annotated on the image with blue and green dots, respectively.
(IF SPEECH CONDITION) 3. **Speech Utterance**: A text transcript of what the user said during the video segment.
4. **Overall Intent**: A summary of what the user aims to achieve in the task segment.

### Outputs:
1. **Task Segment Description**: A very detailed text description of the task segment based on the provided video frames and user intent. Describe the segment to help someone else deeply understand the interaction taking place as if the other user would have to carry out the same interaction from your description. Take into account any segment history provided.
2. **Top-3 Key Frames**: The three most descriptive frames that provide the best understanding of the task segment.
3. **Top-3 Key Frame Description**: A very detailed task-based description of each key frame. Focus on describing details in the frame that would help another user understand the objects being interacted with, how they are interacted with, and how this interaction relates to the user intent. 
4. **Binary Segment Importance**: True of False. If True, the segment is important for understanding the task. If False, the segment is uninformative and should not be examined later to understand the task.

### Guidelines:
- Select the three frames that most effectively illustrate the critical moments or actions in the task.
- If gaze and hand information is annotated on the image, use it to identify what the user is focusing on when asking the query, and what you should focus on when giving your answer.
- Segment importance should be marked as 'False' if the segment shows idle moments, transition phases, redundant actions, errors or missteps, or background activities that are not relevant to understanding the task.
- Be specific about the key frame based on the user's environment, task, and user preferences as much as you are confident. For example, specifying 'black bin' is more informative than saying 'bin'. Be very detailed in your descriptions and how they relate to the user's intent.
- Think of what would be most informative to teach another user the important actions, visual elements, and user preferences required to carry out the task, so that the user could fully understand the segment/key frame and the intent of the segment/key frame from your description.

### JSON Output Structure:
```json
{
  "task_segment_description": "[Your detailed description here]",
  "key_frames": [
    {
      "frame_number": X,
      "reason": "[Description of why this frame is key]"
    },
    {
      "frame_number": Y,
      "reason": "[Description of why this frame is key]"
    },
    {
      "frame_number": Z,
      "reason": "[Description of why this frame is key]"
    }
   ],
  "is_segment_important": [Boolean True/False of whether segment is important]
}
```

\end{lstlisting}

\onecolumn\begin{lstlisting}[caption={Prompt template for inference on new examples},captionpos=t,label={prompt_inference}] 
Task: You are an assistant guiding a user to complete a task exactly the same as an expert has demonstrated to you. The user has provided you with a natural language query and an egocentric view from a head-mounted camera. You also have access to experience data from the expert who has performed the same task. It is important to note that the user asking the query cannot see and has never seen the expert's experience. Your goal is to generate a response that accurately addresses the user's query and helps them perform the task as effectively as the expert. Ensure that your response incorporates the visual context from the camera and leverages the expert's experience to provide a precise and helpful answer.

Inputs:
- Chat History: If provided, chat history between you and the user from this interaction session.
- Expert Experience: You have access to a collection of key frames and their descriptions showing an expert demonstrating the task. Each segment contains several key frames, capturing images and descriptions of the expert performing the task correctly. Remember, only you can see this past experience; the user cannot and has never seen it. Therefore, you must describe the expert's experience in full detail so the user can understand and perform the task accurately.
    - The user's eye gaze location during the query may be annotated on the image with a purple circle. The user's right and left hand locations are annotated on the image with blue and green dots, respectively.
- User Intent: The overall user intent of what task the user is performing. This may help contextualize the query.
- Egocentric View: An egocentric image that is likely important for answering the query.
- Natural Language Query: The user query that you should respond to. The query answer will likely require information found in the egocentric view and the previous experience.

Output:
- Answer: Your answer to the query based on the inputs. Answer the Natural Language Query directly. Be concise without leaving out important information. Your answer should not be more than one or two sentences. You should only respond in English, and should ask for the user to repeat if the query is not in English.

Additional Context and Guidelines:
- You should guide the user so that the user completes the task the same as the expert has demonstrated, without deviating. For example, if the user intent is shopping for X, the same items as the expert should be bought.
- The user is wearing a head-mounted camera that provides the egocentric view to you of their environment.
- The previous experience may be from the distant past and the environment or user state may have changed. Use the previous experience as context given that aspects of the current scene may have changed.
- Format your output as a json, with "answer" as a seperate key.
- Do not refer to the expert experience in your answer without describing the experience fully. The user will not understand utterances such as "like in experience #1" or "refer to expert previous experience".
- If gaze and hand information is annotated on the image, use it to identify what the user is focusing on when asking the query, and what you should focus on when giving your answer.
- Only use the chat history to remember what was just said. Use the expert experience to help you answer the question.
- Be as concise as possible in your answer without leaving out important details.
- If the task involves food items, never instruct the user to consume the food. You can instruct them to prepare the food, buy the food, etc. but not consume or ingest. 
- You should only respond in English. If the Natural Language Query is not in English, you should respond "Sorry I didn't hear that correctly, can you repeat that?"
\end{lstlisting}

\lstset{escapeinside={<@}{@>}, language=}
\onecolumn\begin{lstlisting}[caption={Task instructions for phase 1.},captionpos=t,label={task_instruction_phase1}] 
Phase 1 Instructions
You will be carrying out a specified task while wearing the HoloLens. Please follow these instructions carefully:
1.	Preparation:
o	Ensure the HoloLens is properly fitted and functioning.
o	Familiarize yourself with the task you will be performing.
2.	Starting the Task:
o	Begin by clearly stating aloud the specific task you are about to perform. For example, say, "I am going to make a cup of tea."
o	Once you have stated your task, proceed to start performing it.
3.	During the Task:
o	As you perform the task, continuously speak aloud to describe your actions and intentions. This is known as "thinking aloud." For example, if you are making a cup of tea, you might say, "I am filling the kettle with water," or "I am adding a tea bag to the cup."
o	Be natural in your descriptions as if you were demonstrating the task to someone. This helps to annotate your actions and intentions accurately.
4.	Performing the Task Naturally:
o	Carry out the task as naturally as you would in your daily life. Do not alter your usual way of doing things.
o	If you encounter any issues or have to deviate from your normal process, describe these changes aloud as well.
o	You can refer to items by looking at them or by using hand gestures, such as pointing.
o	Important: Keep the scene area you are looking at in the center of your visor. Parts of the scene located well below your visor will not be visible in the recording, even if you can see them.
5.	Completion:
o	Once you have completed the task, clearly state, "Task complete."
o	Inform the study administrator and wait for confirmation from the administrator before taking off the HoloLens.
6.	Respond to questions
o	After completing your task, you will answer 5 questions on the computer by typing in your response on the keyboard. You should check the question and the images that appear on the screen, and answer the query based on your demonstration. 
o	The images are additional context and come directly from your demonstration. 
o	Please answer them as if someone was asking you clarifying questions to understand the why, what, and how of your performed task.
By following these steps, you will help us collect valuable data for our study. Thank you for your participation!
If you are uncomfortable or tired at any point during the study, please let the administrator know and we will stop the collection, without affecting compensation. 
Your task: Please see the second sheet of paper with your task instructions.
\end{lstlisting}

\onecolumn\begin{lstlisting}[caption={Task instructions for phase 2.},captionpos=t,label={task_instruction_phase2}] 
Instructions Phase 2

You will perform a task by interacting with an AI assistant. Follow these instructions:
1.	Task Overview:
o	You will ask an AI assistant questions about a task that another person has previously performed. This is unrelated to the tasks you have previously performed and will not follow the same patterns.
o	The AI assistant can view what you are seeing just before you ask your question (see #2) 
o	Ask questions as naturally as possible, similar to how you would in a real-life scenario where you need help understanding a new task.
2.	Asking Questions:
o	To ask a question, begin by saying the keyword, "Agent" or "Hey agent".
1.	Important! Ensure you are looking at the specific part of the scene or object you want to ask about before saying "Agent" or "Hey agent". The agent will only be able to see what you were looking at before you say the keyword.
2.	Keep the scene area you are looking at in the center field of view of your visor (if you are looking straight ahead). Parts of the scene located below your visor will not be visible in the recording, even if you can see them.
o	The assistant will acknowledge by responding back, indicating it is ready to listen to your question. If the agent does not respond back, you may need to say the keyword again.
o	Once the assistant has acknowledged, proceed to ask your question.
3.	Example Flow:
o	Example Interaction 1: Morning routine
1.	You: (While looking at different bottles.) "Hey agent." 
2.	Assistant: "How can I assist you?"
3.	You: "Which of these should I start with?"
4.	Assistant: "First, open the red bottle and pour water in it."
5.	You: (You follow the instructions.)
6.	You: (while looking at the bottle with water) "Hey agent" 
7.	Assistant: "Go ahead."
8.	You: Is this correct?
9.  ...
o	Example Interaction 2: Organizing
1.  ... (previous interactions)
2.	You: (While holding an apple in your hand and looking at it.) "Hey agent." 
3.	Assistant: "I'm here. How can I assist you?"
4.	You: "What should I do with this?"
5.	Assistant: "Put that in the bowl on the shelf."
6.	You: (Put the apple in the bowl on the shelf.)
7.  ...
o	Example Interaction 3: Decluttering the Desk
1.  ... (previous interactions)
2.	You: (While looking at a de-cluttered desk.) "Hey agent." 
3.	Assistant: "I'm here. How can I assist you?"
4.	You: "Is this correct?"
5.	Assistant: "No, the pens should be put in the cup on the right."
6.	You: (You put pens in the cup on the right.)
7.  ...
4.	General Tips:
o	Keep your questions clear and specific to the part of the task you need help with.
o	If the assistant's response is unclear, feel free to ask follow-up questions for further clarification.
o	Important! The assistant may give confusing advice. Follow the assistant's responses as much as possible. Try not to get stuck on a part of the task too long. If you do not get the guidance you need, use your best judgement and move to the next part of the task. It is fine if you are not able to complete the task.
o	It is OK if you cannot complete the task successfully. If the agent gives repetitive or unclear advice for multiple rounds, we suggest you move onto a different part of the task, or end the task completely by saying "task complete".
o	It can be beneficial to ask about specific items or groups of specific, rather than very general questions about the task.
o	You can refer to items by looking at them or by using hand gestures, such as pointing.
By following these steps, you will help us collect valuable data for our study. Thank you for your participation!
If you are uncomfortable or tired at any point during the study, please let the administrator know and we will stop the collection, without affecting compensation. 

Your task: Please see the second sheet of paper for your task instructions.
\end{lstlisting}

\end{document}